%% file: paper.tex
\documentclass[runningheads]{llncs}

\input{math_commands.tex}

\usepackage[utf8]{inputenc} 
\usepackage[T1]{fontenc}    
\usepackage{hyperref}       
\usepackage{url}            
\usepackage{booktabs}       
\usepackage{cite}
\usepackage{amsfonts}       
\usepackage{nicefrac}       
\usepackage{microtype}      
\usepackage{pbox}
\usepackage{adjustbox}
\usepackage{appendix}
\usepackage{bm}
\usepackage{graphicx}
\usepackage{booktabs} 
\usepackage{siunitx}
\usepackage{xspace}
\usepackage{xparse}

\usepackage{hyperref}

\usepackage[algo2e,noend,linesnumberedhidden,ruled,resetcount,noline]{algorithm2e}
\usepackage{algorithm}

\usepackage{multirow,verbatim,color}

\usepackage{latexsym,amsmath,amssymb,amsfonts}
\usepackage{setspace}
\usepackage{enumitem}
\usepackage{textcomp}
\usepackage{stmaryrd}
\usepackage{etoolbox}
\usepackage[english]{babel}
\usepackage[normalem]{ulem}
\usepackage{caption}
\usepackage{subcaption}

\usepackage{pgf}
\usepackage{tikz}

\fontfamily{ptm}

\addto\extrasenglish{%
}

\usepackage{etoolbox}
\makeatletter
\patchcmd{\hyper@makecurrent}{%
    \ifx\Hy@param\Hy@chapterstring
        \let\Hy@param\Hy@chapapp
    \fi
}{%
    \iftoggle{inappendix}{
        \@checkappendixparam{chapter}%
        \@checkappendixparam{section}%
        \@checkappendixparam{subsection}%
        \@checkappendixparam{subsubsection}%
        \@checkappendixparam{paragraph}%
        \@checkappendixparam{subparagraph}%
    }{}%
}{}{\errmessage{failed to patch}}

\newcommand*{\@checkappendixparam}[1]{%
    \def\@checkappendixparamtmp{#1}%
    \ifx\Hy@param\@checkappendixparamtmp
        \let\Hy@param\Hy@appendixstring
    \fi
}
\makeatletter

\newtoggle{inappendix}
\togglefalse{inappendix}

\apptocmd{\appendix}{\toggletrue{inappendix}}{}{\errmessage{failed to patch}}
\apptocmd{\subappendices}{\toggletrue{inappendix}}{}{\errmessage{failed to patch}}

\input{defs}

\title{On Relating `Why?' and `Why Not?' Explanations}

\author{
  Alexey Ignatiev\inst{1} \and
  Nina Narodytska\inst{2} \and
  Nicholas Asher\inst{3} \and
  Joao Marques-Silva\inst{3}
}
\authorrunning{Ignatiev et al.}

\institute{
  Monash University, Melbourne, Australia
  \href{mailto:alexey.ignatiev@monash.edu}{\texttt{alexey.ignatiev@monash.edu}} \and
  VMware Research, CA, USA
  \href{mailto: nnarodytska@vmware.com}{\texttt{nnarodytska@vmware.com}} \and
  ANITI, IRIT, CNRS, Toulouse, France \\
  \{\mailtodomain{nicholas.asher}\texttt{,}\mailtodomain{joao.marques-silva}\}\texttt{@irit.fr}
}

\begin{document}

\maketitle

\input{abs}
%
\input{intro}

\input{prelim}

\input{xps}

\input{dual}
\input{algs}

\input{res}
\input{conc}

%
%
\bibliography{refs,xai}
\bibliographystyle{abbrv}
\input{appendix}
\end{document}

%% file: math_commands.tex
\usepackage{amsmath,amsfonts,bm}

\def\eqref#1{equation~\ref{#1}}

\def\1{\bm{1}}

\DeclareMathAlphabet{\mathsfit}{\encodingdefault}{\sfdefault}{m}{sl}
\SetMathAlphabet{\mathsfit}{bold}{\encodingdefault}{\sfdefault}{bx}{n}

\DeclareMathOperator*{\argmin}{arg\,min}

%% file: defs.tex
\newenvironment{Proof}{\noindent{\em Proof.~}}{\hfill$\Box$\\[-0.15cm]}

\newcommand{\todoF}[2]{}

\newcommand{\fml}[1]{{\mathcal{#1}}}

\newcommand{\tn}[1]{\textnormal{#1}}

\newcommand{\tsf}[1]{$\mathsf{#1}$}

\newcommand{\msf}[1]{\mathsf{#1}}

\newcommand{\mbb}[1]{\mathbb{#1}}

\newcommand{\eqlit}[2]{(\ensuremath\msf{#1}=\msf{#2})\xspace}

\DeclareMathOperator*{\nentails}{\nvDash}
\DeclareMathOperator*{\entails}{\vDash}

\DeclareMathOperator*{\limply}{\rightarrow}

\newcommand{\mrel}{\ensuremath{\fml{M}_{\pi}}}
\newcommand{\mrelx}[1]{\ensuremath{\fml{M}_{#1}}}
\newcommand{\tval}{\ensuremath\mathbf{true}}
\newcommand{\fval}{\ensuremath\mathbf{false}}

\NewDocumentCommand{\axpq}{ O{prediction~$\pi$} }{`Why\xspace#1?'\xspace}
\NewDocumentCommand{\cxpq}{ O{pred.~$\pi$} O{pred.~$\neg\pi$}}{`Why\xspace#1\xspace and not\xspace#2?'\xspace}
\NewDocumentCommand{\cxpqp}{ O{pred.~$\pi$} O{pred.~$\neg\pi$}}{`Why\xspace(#1\xspace and) not\xspace#2?'\xspace}
\NewDocumentCommand{\axpdef}{ }{`Why?'\xspace}
\NewDocumentCommand{\cxpdef}{ }{`Why Not?'\xspace}

\definecolor{gray}{rgb}{.4,.4,.4}
\definecolor{midgrey}{rgb}{0.5,0.5,0.5}
\definecolor{middarkgrey}{rgb}{0.35,0.35,0.35}
\definecolor{darkgrey}{rgb}{0.3,0.3,0.3}
\definecolor{darkred}{rgb}{0.7,0.1,0.1}
\definecolor{midblue}{rgb}{0.2,0.2,0.7}
\definecolor{darkblue}{rgb}{0.1,0.1,0.5}
\definecolor{defseagreen}{cmyk}{0.69,0,0.50,0}

\newcommand{\jnoteF}[1]{}
\newcommand{\anoteF}[1]{}
\newcommand{\nnoteF}[1]{}

\newcounter{Comment}[Comment]
\setlist{nosep}
\setitemize{itemsep=0pt,topsep=0pt,partopsep=0pt,parsep=0pt}
\setenumerate{itemsep=0pt,topsep=0pt,partopsep=0pt,parsep=0pt}

\newcommand{\zerodisplayskips}{%
  \setlength{\abovedisplayskip}{4.0pt}%
  \setlength{\belowdisplayskip}{4.0pt}%
  \setlength{\abovedisplayshortskip}{4.0pt}%
  \setlength{\belowdisplayshortskip}{4.0pt}}
\appto{\normalsize}{\zerodisplayskips}
\appto{\small}{\zerodisplayskips}
\appto{\footnotesize}{\zerodisplayskips}

\input{./algs/alg-defs}

\SetAlCapSty{}
\SetAlCapSkip{0.5em}
\newcommand{\TopBlankLine}{\vspace*{3.5pt}}
\newcommand{\BotBlankLine}{\vspace*{1.5pt}}

\newcommand{\mailtodomain}[1]{\href{mailto:#1@irit.fr}{\texttt{\nolinkurl{#1}}}}


%% file: algs/alg-defs.tex
\SetAlgoNoEnd
\SetAlgoNoLine
\SetFillComment
\SetKwBlock{Let}{let}{}
\SetKwBlock{FBlock}{}{}
\SetKw{KwNot}{not\xspace}
\SetKw{KwAnd}{and\xspace}
\SetKw{KwOr}{or\xspace}
\SetKw{Break}{break\xspace}
\SetKwData{false}{{\small false}}
\SetKwData{true}{{\small true}}
\SetKwData{st}{\small{\sl st}}
\SetKwFunction{SAT}{SAT}
\SetKwFunction{Entails}{Entails}
\SetKwFunction{FindMHS}{FindMHS}
\SetKwFunction{Clauses}{Clausify}
\SetKwFunction{ExtractAXp}{ExtractAXp}
\SetKwFunction{ReportAXp}{ReportAXp}
\SetKwFunction{ExtractCXp}{ExtractCXp}
\SetKwFunction{ReportCXp}{ReportCXp}
\SetKwFunction{NegLits}{NegateLiteralsOf}
\SetKwFunction{PosLits}{UseLiteralsOf}
\SetKwBlock{Let}{let}{end}
\SetKwBlock{FBlock}{}{end}
\SetKwInput{KwVars}{Variables}
\SetKwInOut{Global}{Global}
\SetKwHangingKw{Algorithm}{Algorithm}
\SetKwHangingKw{Function}{function}
\SetKw{Func}{Function}
\SetKwBlock{Begin}{}{}

%% file: abs.tex
\begin{abstract}
  Explanations of Machine Learning (ML) models often address a
  \axpdef question.
  Such explanations can be related with selecting feature-value pairs
  which are sufficient for the prediction.
  Recent work has investigated explanations that address a \cxpdef
  question, i.e.\ finding a change of feature values that guarantee a
  change of prediction. 
  Given their goals, these two forms of explaining predictions of ML
  models appear to be mostly unrelated.
  However, this paper demonstrates otherwise, and establishes a
  rigorous formal relationship between \axpdef and \cxpdef
  explanations.
  Concretely, the paper proves that, for any given instance,
  \axpdef explanations are minimal hitting sets of
  \cxpdef explanations and vice-versa.
  Furthermore, the paper 
  devises novel algorithms for extracting and enumerating both forms
  of explanations.
\end{abstract}

%% file: intro.tex
\section{Introduction} \label{sec:intro}

The importance of devising mechanisms for computing explanations of
Machine Learning (ML) models cannot be overstated, as illustrated by
the fast-growing body of work in this area.
A glimpse of the importance of explainable AI (XAI) is offered by a
growing number of recent surveys and overviews~\cite{klein-ieee-is17a,klein-ieee-is17b,cotton-ijcai07-xai,muller-dsp18,klein-ieee-is18a,klein-ieee-is18b,berrada-ieee-access18,mencar-ipmu18,hlupic-mipro18,klein-corr18,pedreschi-acmcs19,xai-bk19,muller-xai19-ch01,miller-aij19,miller-acm-xrds19,anjomshoae-aamas19,russell-fat19a,zhu-nlpcc19}.

Past work on computing explanations has mostly addressed \emph{local}
(or instance-dependent)
explanations~\cite{guestrin-kdd16,lundberg-nips17,guestrin-aaai18,darwiche-ijcai18,darwiche-aaai19,inms-aaai19,darwiche-ecai20,darwiche-pods20}. 
%
Exceptions include for example approaches that distill ML models,
e.g.\ the case of NNs~\cite{hinton-cex17} among many
others~\cite{guestrin-kdd16}, or recent work on relating explanations
with adversarial examples~\cite{inms-nips19}, both of which can be
seen as seeking \emph{global} (or instance-independent) explanations.
Prior research has also mostly considered model-agnostic
explanations~\cite{guestrin-kdd16,lundberg-nips17,guestrin-aaai18}.
%
Recent work on model-based explanations, e.g.~\cite{darwiche-ijcai18,inms-aaai19}, refers to local (or global)
model-agnostic
explanations as \emph{heuristic}, given that these
approaches offer no \emph{formal} guarantees with respect to the
underlying ML model\footnote{A taxonomy of ML model explanations used
  in this paper is included in~\autoref{app:taxonomy}.}.
Examples of heuristic approaches
include~\cite{guestrin-kdd16,lundberg-nips17,guestrin-aaai18}, among
many others\footnote{%
  There is also a recent XAI service offered by Google:
  \url{https://cloud.google.com/explainable-ai/},
  inspired on similar ideas~\cite{google-xai-whitepaper}.}.
In contrast, local (or global) model-based explanations are referred
to as \emph{rigorous}, since these offer the strongest formal
guarantees with respect to the underlying ML model. Concrete examples
of such rigorous approaches
include~\cite{darwiche-ijcai18,TranG18,darwiche-aaai19,inms-aaai19,inms-corr19,nsmims-sat19,inms-nips19,darwiche-ecai20,darwiche-pods20,JhaSRPF19,msgcin-nips20,ignatiev-ijcai20,iims-corr20}.

Most work on computing explanations aims to answer a
\axpq question.
Some work proposes approximating the ML model's behavior with a linear
model~\cite{guestrin-kdd16,lundberg-nips17}. Most other work seeks to
find a (often minimal) set of feature value pairs which is sufficient
for the prediction, i.e.\ as long as those features take the specified
values, the prediction does not change.
For rigorous approaches, the answer to a \axpq
question has been referred to as
PI-explanations~\cite{darwiche-ijcai18,darwiche-aaai19}, abductive
explanations~\cite{inms-aaai19}, but also as (minimal) sufficient
reasons~\cite{darwiche-ecai20,darwiche-pods20}. (Hereinafter,
we 
use the term \emph{abductive explanation} because of
the other forms of explanations studied in the paper.)

Another dimension of explanations, studied in recent
work~\cite{miller-aij19}, is the  difference between explanations for
\axpq
questions, e.g.,  `Why did I get the loan?', and for
\cxpq[prediction~$\pi$][$\delta$]
questions, e.g., `Why didn't I get the loan?'.
Explanations for \cxpdef
questions, labelled by \cite{miller-aij19} \emph{contrastive}
explanations, isolate a pragmatic component of explanations that
\emph{abductive explanations} lack.
Concretely, an abductive explanation identifies a set of feature
values which are sufficient for the model to make a prediction $\pi$
and thus provides an answer to the question
\axpq[$\pi$]
A constrastive explanation sets up a counterfactual link between what
was a
(possibly) {\em desired} outcome of a certain set of features and what
was the observed outcome \cite{bromberger:1962,achinstein:1980}.  Thus, a
contrastive explanation answers a
\cxpq[$\pi$][$\delta$]
question
\cite{miller-corr18,das-nips18,russell-fat19a}.
%
%
%
%

In this paper we focus on the relationship between \emph{local}
abductive and contrastive explanations\footnote{In contrast with recent work~\cite{inms-nips19}, which studies the
relationship between \emph{global} model-based (abductive)
explanations and adversarial examples.}. One of our 
contributions is to show how recent approaches for computing rigorous
abductive
explanations~\cite{darwiche-ijcai18,darwiche-aaai19,inms-aaai19,darwiche-ecai20,darwiche-pods20}
can also be exploited for computing contrastive explanations.
To our knowledge, this
is new.  In addition,
we demonstrate that rigorous (model-based) local
abductive and contrastive explanations are related by a minimal
hitting set relationship~\footnote{A local
abductive (resp. contrastive)  explanation is a minimal
hitting set of the set of all local contrastive (resp.
abductive) explanations.}, which builds on the seminal work of Reiter
in the 80s~\cite{reiter-aij87}.
Crucially, this novel hitting set relationship reveals a
wealth of algorithms for computing and for enumerating contrastive and
abductive explanations. We emphasize that it 
allows designing the first
algorithm to \emph{enumerate} abductive explanations.
Finally, we  demonstrate feasibility of our approach experimentally.
Furthermore, our experiments show that there is a strong correlation
between contrastive explanations and explanations produced by the commonly used SHAP explainer.

%% file: prelim.tex
\section{Preliminaries} \label{sec:prelim}

\paragraph{Explainability in Machine Learning.}
The paper assumes an ML model $\mathbb{M}$, which is represented by
a finite set of first-order logic (FOL) sentences $\fml{M}$. (When
applicable, simpler alternative representations for $\fml{M}$ can be
considered, e.g.\ (decidable) fragments of FOL, (mix\-ed-)integer
linear programming, constraint language(s), etc.)%
\footnote{%
  $\fml{M}$ is referred to as the (formal) model of the ML model
  $\mbb{M}$. The use of FOL is not restrictive, with fragments of FOL
  being used in recent years for modeling ML models in different
  settings. These include NNs~\cite{inms-aaai19} and
  Bayesian Network Classifiers~\cite{darwiche-aaai19}, among others.}
A set of features $\fml{F}=\{f_1,\ldots,f_L\}$ is assumed.
Each feature $f_i$ is categorical (or ordinal), with values taken from
some set $D_i$. An \emph{instance} is an assignment of values to
features.
The space of instances, also referred to as \emph{feature} (or
\emph{instance}) \emph{space}, is defined by
$\mathbb{F}={D_1}\times{D_2}\times\ldots\times{D_L}$.
(
For real-valued features, a suitable interval discretization can be
considered.)
A (feature) literal $\lambda_i$ is of the form $(f_i=v_i)$, with
$v_i\in{D_i}$. In what follows, a literal will be viewed as an atom,
i.e.\ it can take value \emph{true} or \emph{false}.
As a result, an instance can be viewed as a set of $L$ literals,
denoting the $L$ distinct features, i.e.\ an instance contains a
single occurrence of a literal defined on any given feature.
A set of literals is consistent if it contains at most one literal
defined on each feature.
A consistent set of literals can be interpreted as a conjunction or as
a disjunction of literals; this will be clear from the context. When
interpreted as a conjunction, the set of literals denotes a
\emph{cube} in instance space, where the unspecified features can take
any possible value of their domain. When interpreted as a disjunction,
the set of literals denotes a \emph{clause} in instance space. As
before, the unspecified features can take any possible value of their
domain.

The remainder of the paper assumes a classification problem with a set
of classes $\mathbb{K}=\{\kappa_1,\ldots,\kappa_M\}$. A prediction
$\pi\in\mathbb{K}$ is associated with each instance
$X \in\mbb{F}$.
Throughout this paper, an ML model $\mbb{M}$ will be associated with
some logical representation (or encoding), whose consistency depends
on the (input) instance and (output) prediction.
Thus, we define a predicate $\fml{M}\subseteq\mbb{F}\times\mbb{K}$,
such that $\fml{M}(X,\pi)$ is
true iff the input $X$ is consistent with prediction $\pi$ given the
ML model $\mbb{M}$\footnote{%
  This alternative notation is used for simplicity and clarity
  with respect to earlier
  work~\cite{darwiche-ijcai18,inms-aaai19,inms-nips19}.
  %
  %
  Furthermore, defining $\fml{M}$ as a predicate allows for multiple
  predictions for the same point in feature space. Nevertheless, such
  cases are not considered in this paper.}.
We further simplify the notation by using $\fml{M}_{\pi}(X)$ to denote
a predicate $\fml{M}(X,\pi)$ for a concrete prediction $\pi$.

%
Moreover, we  will compute \emph{prime implicants}
of $\mrel$. These 
predicates defined on $\mbb{F}$ and represented as consistent
conjunctions (or alternatively as sets) of feature literals.
Concretely, a consistent conjunction of feature literals $\tau$ is an
implicant of $\mrel$ if the following FOL statement is true:
\begin{equation}
  \forall(X\in\mathbb{F}).\tau(X)\limply\fml{M}(X,\pi)
\end{equation}
The notation $\tau\entails\mrel$ is used to denote that $\tau$ an
implicant of $\mrel$.
%
Similarly, a consistent set of feature literals $\nu$ is the
negation of an implicate of $\mrel$ if the following FOL statement is
true:
\begin{equation}
  \forall(X\in\mathbb{F}).\nu(X)\limply\left(\lor_{\rho\not=\pi}\fml{M}(X,\rho)\right)
\end{equation}
$\mrel\entails\neg\nu$, or
alternatively
$\left(\nu\entails\neg\mrel\right)\equiv\left(\nu\entails\lor_{\rho\not=\pi}\mrelx{\rho}\right)$.
An implicant $\tau$ (resp.\ implicate $\nu$) is called \emph{prime} if
none of its proper subsets $\tau'\subsetneq\tau$
(resp.\ $\nu'\subsetneq\nu$) is an implicant (resp.\ implicate).

Abductive explanations represent prime implicants of the decision
function associated with some predicted class $\pi$\footnote{%
  By definition of prime implicant, abductive explanations are
  sufficient reasons for the prediction. Hence the names used in
  recent work: abductive explanations~\cite{inms-aaai19},
  PI-explanations~\cite{darwiche-ijcai18,darwiche-aaai19} and
  sufficient reasons~\cite{darwiche-ecai20,darwiche-pods20}.}.

\textbf{Analysis of Inconsistent Formulas.} 
Throughout the paper, we will be interested in 
formulas $\fml{F}$ that are \emph{inconsistent} (or
\emph{unsatisfiable}), i.e.\ $\fml{F}\entails\bot$, represented as
conjunctions of clauses. Some clauses in $\fml{F}$ can be {\em
  relaxed} (i.e.\ allowed not to be satisfied) to restore consistency,
whereas others cannot.
Thus, we assume that $\fml{F}$ is partitioned into two first-order
subformulas $\fml{F}=\fml{B}\cup\fml{R}$, where $\fml{R}$ contains the
\emph{relaxable} clauses, and $\fml{B}$ contains the \emph{non-relaxable}
clauses. $\fml{B}$ can be viewed as (consistent) background knowledge,
which must always be satisfied.

Given an inconsistent formula $\fml{F}$, represented as a set of
first-order clauses, we identify
the clauses that are
responsible for unsatisfiability among those that can be relaxed, as
defined next\footnote{%
  The definitions in this section are often presented for the
  propositional case, but the extension to the first-order case is
  straightforward.}.
\begin{definition}[Minimal Unsatisfiable Subset (MUS)] \label{def:mus}
  Let $\fml{F}=\fml{B}\cup\fml{R}$ denote an inconsistent set of
  clauses ($\fml{F}\entails\bot$).
  $\fml{U}\subseteq\fml{R}$ is a {\em  Minimal Unsatisfiable Subset}
  (MUS) iff $\fml{B}\cup\fml{U}\entails\bot$ and
  $\forall_{\fml{U}'\subsetneq\fml{U}},\,\fml{B}\cup\fml{U}'\nentails\bot$.
  %
\end{definition}
Informally, an MUS provides the minimal information that needs to be
added to the background knowledge $\fml{B}$ to obtain an
inconsistency; 
it explains the causes for this
inconsistency. Alternatively, one might be interested in correcting
the formula, removing some clauses in $\fml{R}$ to achieve
consistency.
\begin{definition}[Minimal Correction Subset (MCS)] \label{def:mcs}  
  Let $\fml{F}=\fml{B}\cup\fml{R}$ denote an inconsistent set of
  clauses ($\fml{F}\entails\bot$).
  $\fml{T}\subseteq\fml{R}$ is a {\em Minimal Correction Subset} (MCS)
  iff $\fml{B}\cup\fml{R}\setminus\fml{T}\nentails\bot$ and
  $\forall_{\fml{T}'\subsetneq\fml{T}}$, $\fml{B}\cup\fml{R}\setminus\fml{T}'\entails\bot$.
\end{definition}
A fundamental result in reasoning about inconsistent clause sets is
the minimal hitting set (MHS) duality relationship between MUSes and
MCSes~\cite{reiter-aij87,lozinskii-jetai03}:
\emph{MCSes are MHSes of MUSes and vice-versa.}
This result has been extensively used in the development of
algorithms for MUSes and MCSes 
~\cite{stuckey-padl05,liffiton-jar08,lpmms-cj16}, and also applied
in a number of different settings.
Recent years have witnessed the proposal of a large number of novel
algorithms for the extraction and enumeration of MUSes and
MCSes~\cite{bacchus-cav15,lpmms-cj16,lagniez-ijcai18,bendik-atva18}.
%
Although most work addresses propositional theories, these
algorithms can easily
be generalized to any other setting where
entailment is monotonic, e.g.\ SMT~\cite{demoura-tacas08}.

\begin{figure*}[t]
  \begin{minipage}{0.3125\linewidth}
    \subcaptionbox{Decision~tree\label{fig:ex01-dt}}[0.9975\linewidth][c]{\begin{center}\input{./texfigs/tree-reads}\end{center}}
  \end{minipage}
  \begin{minipage}{0.68725\linewidth}
    \subcaptionbox{Rule~set\label{fig:ex01-ds}}[0.9975\linewidth][c]{\input{./texfigs/ruleset}}

    \subcaptionbox{Encoding of
      $\fml{M}_{\pi}$\label{fig:ex01-enc}}[0.9975\linewidth][c]{\input{./texfigs/encoding1}}
  \end{minipage}
  \caption{Running example~\cite{poole-bk10}} \label{fig:ex01}
\end{figure*}
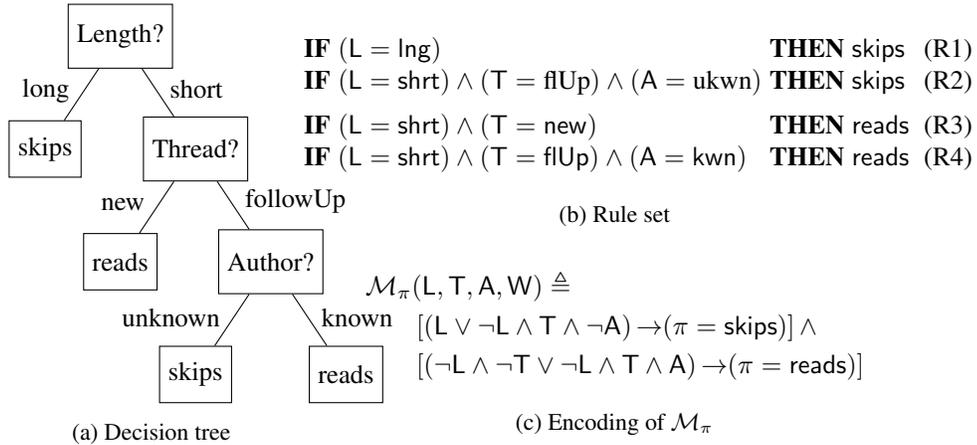

\textbf{Running Example.}
The following example will be used
to illustrate the main ideas.%
%
\begin{example} \label{ex:def1a}
  We consider a textbook
  example~\cite{poole-bk10}[Figure~7.1,~page~289] addressing the
  classification of a user's preferences regarding whether to read or
  to skip a given book.
  For this dataset, the set of features is:
  \[
  \begin{array}{l}
    \{~\msf{A(uthor)},\msf{T(hread)},\msf{L(ength)},\msf{W(hereRead)}~\}\\
  \end{array}\]
  All features take one of two values, respectively
  $\{\msf{known},\msf{unknown}\}$,
  $\{\msf{new},\msf{followUp}\}$,
  $\{\msf{long},\msf{short}\}$, and
  $\{\msf{home},\msf{work}\}$.
  An example instance is:
  $\{
  (\msf{A}=\msf{known}),
  (\msf{T}=\msf{new}),$
  $(\msf{L}=\msf{long}),
  (\msf{W}=\msf{home}) \}$.
  This instance is identified as $e_1$~\cite{poole-bk10} with
  prediction \tsf{skips}.
  \autoref{fig:ex01-dt} shows a possible decision tree
  for this example~\cite{poole-bk10}%
  \footnote{%
    The choice of a decision tree aims only at keeping the example(s)
    presented in the paper as simple as possible.
    The ideas proposed in the paper apply to \emph{any} ML model that
    can be represented with FOL. This encompasses \emph{any} existing
    ML model, with minor adaptations in case the ML model keeps state.}.
  The decision tree can 
  be represented as a set of rules as shown
  in~\autoref{fig:ex01-ds}\footnote{%
    The abbreviations used relate with the
    names in the decision tree, and serve for saving space.}.
\end{example}

Our goal is to reason about the ML model, i.e.\ to implement
model-based reasoning, so we need to propose a logical representation for the ML model.


\begin{example}  \label{ex:def01b}
  For implementing model-based reasoning, we need to develop an
  encoding in some suitable fragment of FOL~\footnote{%
    Depending on the ML problem, more expressive fragments of FOL
    logic could be considered~\cite{kroening-bk16}. Well-known
    examples include real, integer and integer-real arithmetic, but
    also nonlinear arithmetic~\cite{kroening-bk16}.}. 0-place 
  predicates~\footnote{%
    Which in this case are used as propositional variables.}
  are used for $\msf{L}$, $\msf{T}$, $\msf{A}$ and $\msf{W}$, as
  follows. We will associate $(\msf{L}=\msf{long})$ with $\msf{L}$
  and $(\msf{L}=\msf{short})$ with $\neg\msf{L}$. Similarly, we
  associate $(\msf{T}=\msf{new})$ with $\neg\msf{T}$, and
  $(\msf{T}=\msf{followUp})$ with $\msf{T}$.
  We associate $(\msf{A}=\msf{known})$ with $\msf{A}$ and
  $(\msf{A}=\msf{unknown})$ with $\neg\msf{A}$.
  Furthermore, we associate $(\msf{W}=\msf{home})$ with $\neg\msf{W}$
  and $(\msf{W}=\msf{work})$ with $\msf{W}$.
  %
  %
  An example encoding is shown in~\autoref{fig:ex01-enc}. The explicit
  values of $\pi$ are optional (i.e.\ propositional values could be
  used) and serve to illustrate how non-propositional valued could be
  modeled.
\end{example}

%% file: texfigs/tree-reads.tex
%
\tikzset
{
  inode/.style = {rectangle, draw=black, align=center, minimum size=0.85cm},
  lnode/.style  = {rectangle, draw=black, align=center, minimum height=0.75cm, minimum width=0.75cm}
}
\begin{tikzpicture}[level distance=1.5cm,
    level 1/.style={sibling distance=2.0cm},
    level 2/.style={sibling distance=2.0cm}]

  \node (Root) [inode] {Length?}
  child { node [lnode] {skips} edge from parent node[left,draw=none,xshift=-1pt,yshift=1.5pt] {long} }
  child { node [inode] {Thread?}
    child { node [lnode] {reads} edge from parent node[left,draw=none,xshift=-1pt,yshift=1pt] {new} }
    child { node [inode] {Author?}
      child { node [lnode] {skips} edge from parent node[left,draw=none,xshift=-1pt,yshift=1pt] {unknown} }
      child { node [lnode] {reads} edge from parent node[right,draw=none,xshift=1pt,yshift=1pt] {known} }
      edge from parent node [right,draw=none,xshift=1pt,yshift=1.5pt] {followUp}
    }
    edge from parent node [right,draw=none,xshift=1pt,yshift=1.5pt] {short}
  }
  ;
\end{tikzpicture}
%

%% file: texfigs/ruleset.tex
~
\vspace*{0.15cm}
%
\begin{equation*}
  \begin{array}{llllr}
    \tn{\bf IF}&
    \eqlit{L}{lng}&
    \tn{\bf THEN}&
    \msf{skips}&
    \;\tn{(R1)}
    \\
    \tn{\bf IF}&
    \eqlit{L}{shrt}\land\eqlit{T}{\tn{flUp}}\land\eqlit{A}{\tn{ukwn}}&
    \tn{\bf THEN}&
    \msf{skips}&
    \;\tn{(R2)}
    \\[5pt]
    \tn{\bf IF}&
    \eqlit{L}{shrt}\land\eqlit{T}{new}&
    \tn{\bf THEN}&
    \msf{reads}&
    \;\tn{(R3)}
    \\
    \tn{\bf IF}&
    \eqlit{L}{shrt}\land\eqlit{T}{flUp}\land\eqlit{A}{kwn}&
    \tn{\bf THEN}&
    \msf{reads}&
    \;\tn{(R4)}
    %
  \end{array}
\end{equation*}

%% file: texfigs/encoding1.tex
~
\vspace*{0.25cm}
%
  \begin{align*}
    \fml{M}_{\pi}&(\msf{L},\msf{T},\msf{A},\msf{W}) \triangleq  \\
    &\left[(\msf{L}\lor\neg\msf{L}\land\msf{T}\land\neg\msf{A})\limply(\pi=\msf{skips})\right] 
    \land \\
    & \left[(\neg\msf{L}\land\neg\msf{T}\lor\neg\msf{L}\land\msf{T}\land\msf{A})\limply(\pi=\msf{reads})\right]
  \end{align*}

%% file: xps.tex
\section{Contrastive vs.~Abductive Explanations}  \label{sec:xps}

Recent
work~\cite{darwiche-ijcai18,darwiche-aaai19,inms-aaai19,darwiche-pods20} 
proposed to relate model-based explanations with prime implicants.
All these approaches compute a set of feature values which, if
unchanged, are sufficient for the prediction.
Thus, one can view such explanations as answering a
\axpq[]
question: \emph{the prediction is the one given, as long as some
  selected set of feature values is the one given}.
In this paper, such explanations will be referred to as
\emph{abductive explanations}, motivated by one of the approaches used
for their computation~\cite{inms-aaai19}.

\subsection{Defining Abductive Explanations (AXps)}

As indicated earlier in the paper, we focus on \emph{local
  model-based} explanations.
\begin{definition}[Abductive Explanation] \label{def:axp}
  Given an instance $\tau$, with a prediction $\pi$, and an ML model
  represented with a predicate $\fml{M}_{\pi}$,
  i.e.\ $\tau\entails\fml{M}_{\pi}$, an \emph{abductive explanation}
  is a minimal subset of literals of $\tau$, $\sigma\subseteq\tau$,
  such that $\sigma\entails\fml{M}_{\pi}$.
\end{definition}
\begin{example} \label{ex:axpdef}
  With respect to~\autoref{ex:def1a}, let us consider the instance
  $(\msf{A}=\msf{known},\msf{T}=\msf{new},\msf{L}=\msf{short},\msf{W}=\msf{work})$,
  which we will represent instead as
  $(\msf{A},\neg\msf{T},\neg\msf{L},\msf{W})$,
  corresponding to prediction $\pi=\msf{reads}$.
  By inspection of the decision tree (see\autoref{fig:ex01-dt}), a
  possible answer to the
  \axpq[pred.~$\msf{reads}$]
  question is:
  $\{\neg\msf{L},\neg\msf{T}\}$. In this concrete case we can
  conclude that this is the only abductive explanation, again by
  inspection of the decision tree.
\end{example}

\subsection{Defining Contrastive Explanations (CXps)}

As \cite{miller-aij19} notes, contrastive explanations are,
\begin{enumerate}[topsep=0pt,itemsep=0pt,partopsep=0pt,parsep=0pt]
\item[]
\emph{``sought in response to particular counterfactual cases... That is,
  people do not ask why event $P$ happened,but rather why event $P$
  happened instead of some event $Q$.''}
\end{enumerate}
%
%

As a result, we are interested in providing an answer to the question
\cxpq[$\pi$][$\delta$],
where $\pi$ is the prediction given some instance $\tau$, and $\delta$
is some other (desired) prediction.
\begin{example} \label{ex:cxpdef}
  We consider again ~\autoref{ex:def1a}, but with the instance
  specified in~\autoref{ex:axpdef}.
  A possible answer to the question
  \cxpq[pred.~$\msf{reads}$][pred.~$\msf{skips}$?]
  is $\{\msf{L}\}$. Indeed, given the input instance
  $(\msf{A},\neg\msf{T},\neg\msf{L},\msf{W})$, if the value of feature
  $\msf{L}$ changes from $\msf{short}$ to $\msf{long}$, and the value
  of the other features remains unchanged, then the prediction will
  change from $\msf{reads}$ to $\msf{skips}$.
\end{example}
%

The following definition of a (local model-based) contrastive
explanation captures the intuitive notion of the contrastive explanation
discussed in the example above.
\begin{definition}[Contrastive Explanation] \label{def:cxp}
  Given an instance $\tau$, with a prediction $\pi$, and an ML model
  represented by a predicate $\fml{M}_{\pi}$,
  i.e.\ $\tau\entails\fml{M}_{\pi}$, a \emph{contrastive
    explanation} is a minimal subset of literals of $\tau$,
  $\rho\subseteq\tau$, such that
  $\tau\setminus\rho\nentails\fml{M}_{\pi}$.
\end{definition}
This definition means that, there is an assignment to the features with literals in $\rho$, such that the prediction differs from $\pi$.
%
Observe that a CXp is defined to answer the following (more specific)
question
\cxpqp[pred.~$\pi$][$\neg\pi$].
The more general case of answering the question
\cxpqp[pred.~$\pi$][$\delta$]
will be analyzed later.

%% file: dual.tex
\subsection{Relating Abductive \& Contrastive Explanations}  \label{sec:dual}

The previous section proposed a rigorous, model-based, definition of
contrastive explanation.
Given this definition, one can think of developing dedicated algorithms
that compute CXps using a decision procedure for the logic used for
representing the ML model.
Instead, we adopt a simpler approach. We build on a fundamental result
from model-based diagnosis~\cite{reiter-aij87} (and more generally for
reasoning about inconsistency~\cite{lozinskii-jetai03,stuckey-padl05})
and demonstrate a similar relationship between AXps and CXps. In
turn, this result reveals a variety of novel algorithms for computing CXps,
but also offers ways for enumerating both CXps and AXps.

\textbf{Local Abductive Explanations (AXps).}
Consider a set of feature values $\tau$, s.t. 
the predicion
is $\pi$, for which the notation $\tau\entails\mrel$ is used.
We will use the equivalent statement,
$\tau\land\neg\fml{M}_{\pi}\entails\bot$.
Thus,
\begin{equation} \label{eq:xps}
  \tau\land\neg\fml{M}_{\pi}
\end{equation}
is inconsistent, with the background knowledge being
$\fml{B}\triangleq\neg\fml{M}_{\pi}$ and the relaxable clauses
being $\fml{R}\triangleq\tau$.
As proposed in
~\cite{darwiche-ijcai18,inms-aaai19},
a (local abductive) explanation is a subset-minimal set $\sigma$ of
the literals in $\tau$, such that,
$\sigma\land\neg\fml{M}_{\pi}\entails\bot$. Thus, $\sigma$
denotes a subset of the example's input features which, no matter the
other feature values, ensure that the ML model predicts $\pi$.
Thus, any MUS of~\eqref{eq:xps} is a (local abductive) explanation
for $\mbb{M}$ to predict $\pi$ given $\tau$.

\begin{proposition} \label{prop:axp}
  Local model-based abductive explanations are MUSes of the pair
  $(\fml{B},\fml{R})$,
  $\tau\land\neg\fml{M}_{\pi}$,
  where $\fml{R}\triangleq\tau$ and
  $\fml{B}\triangleq\neg\fml{M}_{\pi}$.
\end{proposition}

\begin{example} \label{ex:lae01}
  Consider the ML model from~\autoref{ex:def1a}, the encoding
  from~\autoref{ex:def01b}, and the instance
  $\{\msf{A},\neg\msf{T},\msf{L},\neg\msf{W}\}$, with prediction
  $\pi=\msf{skips}$ (wrt~\autoref{fig:ex01}, we replace
  $\msf{skips}=\msf{skips}$ with $\tval$ and $\msf{skips}=\msf{reads}$
  with $\fval$).
  We can thus confirm that $\tau\entails\fml{M}_{\pi}$. 
  We observe that the following holds:
  \begin{equation} \label{eq:ex01-xps1}
    \msf{A}\land\neg\msf{T}\land\msf{L}\land\neg\msf{W}
    \entails
    \left[
      \begin{array}{rcl}
        (\msf{L}\lor\neg\msf{L}\land\msf{T}\land\neg\msf{A})&\limply&\tval \\
        & \land & \\
        (\neg\msf{L}\land\neg\msf{T}\lor\neg\msf{L}\land\msf{T}\land\msf{A})&\limply&\fval \\
      \end{array}
    \right]
  \end{equation}
  which can be rewritten as,
  \begin{equation} \label{eq:ex01-xps2}
    \msf{A}\land\neg\msf{T}\land\msf{L}\land\neg\msf{W}
    \land 
    \left[
      \begin{array}{rcl}
        (\msf{L}\lor\neg\msf{L}\land\msf{T}\land\neg\msf{A}) & \land & \neg\tval \\
        & \lor & \\
        (\neg\msf{L}\land\neg\msf{T}\lor\neg\msf{L}\land\msf{T}\land\msf{A}) & \land & \neg\fval \\
      \end{array}
    \right]
  \end{equation}
  It is easy to conclude that ~\eqref{eq:ex01-xps2} is inconsistent.
  %
Moreover,
$\sigma=(\msf{L})$ denotes an MUS of~\eqref{eq:ex01-xps2}
and denotes one abductive explanation for why the prediction is
$\msf{skips}$ for the instance $\tau$.
\end{example}

\textbf{Local Contrastive Explanations (CXps).}
Suppose we compute instead an MCS $\rho$ of~\eqref{eq:xps}, with
$\rho\subseteq\tau$.
As a result,
$\bigwedge_{l\in\tau\setminus\rho}(l)\land\neg\fml{M}_{\pi}\nentails\bot$
holds. Hence, assigning feature values to the inputs of the ML model
is consistent with a prediction that is \emph{not} $\pi$, i.e.\ a
prediction of some value other than $\pi$. Observe that
$\rho$ is a subset-minimal set of literals
which causes $\tau\setminus\rho\land\neg\fml{M}_{\pi}$ to be
satisfiable, with any satisfying assignment yielding a prediction that
is not $\pi$.

\begin{proposition} \label{prop:cxp}
  Local model-based contrastive explanations are MCSes of the pair
  $(\fml{B},\fml{R})$,
  $\tau\land\neg\fml{M}_{\pi}$,
  where $\fml{R}\triangleq\tau$ and
  $\fml{B}\triangleq\neg\fml{M}_{\pi}$.
\end{proposition}

\begin{example} \label{ex:lce01}
  From~\eqref{eq:xps} and \eqref{eq:ex01-xps2} we can also compute
  $\rho\subseteq\tau$ such that
  $\tau\setminus\rho\land\neg\fml{M}_{\pi}\nentails\bot$.
  For example $\rho=(\msf{L})$ is an MCS
  of~\eqref{eq:ex01-xps2}~\footnote{%
    Although in general not the case, in~\autoref{ex:lae01}
    and~\autoref{ex:lce01} an MUS of size  1 is also an MCS of size
    1.}.
  Thus, from $\{\msf{A},\neg\msf{T},\neg\msf{W}\}$ we can get a
  prediction other than $\msf{skips}$, by considering feature value
  $\neg\msf{L}$.
\end{example}

\paragraph{Duality Among Explanations.}
Given the results above, and the hitting set duality
between MUSes and MCSes~\cite{reiter-aij87,lozinskii-jetai03}, we have
the following.
\begin{theorem} \label{th:dual}
  AXps are MHSes of CXps and vice-versa.
\end{theorem}

\begin{Proof}
  Immediate
  from~\autoref{def:axp},~\autoref{def:cxp},~\autoref{prop:axp},~\autoref{prop:cxp},
  and Theorem~4.4 and Corollary~4.5 of~\cite{reiter-aij87}.
\end{Proof}

\autoref{prop:axp},~\autoref{prop:cxp}, and~\autoref{th:dual} can now
serve to exploit the vast body of work on the analysis of inconsistent
formulas for computing both contrastive and abductive explanations
and, arguably more importantly, to enumerate explanations.
%
%
Existing algorithms for the extraction and enumeration of MUSes and
MCSes require minor modications to be applied in the setting of AXps
and CXps (The resulting algorithms are briefly summarized
in~\autoref{app:sec:algs}.
Interestingly, a consequence of the duality is that  computing an abductive explanation is 
\emph{harder} than computing a contrastive explanation in terms of the number of calls to a decision procedure~\autoref{app:sec:algs}.).
\vspace{-0.5em}
\paragraph{Discussion.}
As observed above, the contrastive explanations we are computing
answer the question:
\cxpqp[$\pi$][$\neg\pi$].
A more general contrastive explanation would be
\cxpqp[$\pi$][$\delta$, with $\pi\not=\delta$]~\cite{miller-aij19}.
Note that, since the prediction $\pi$ is given,
we are only interested in changing the prediction to either $\neg\pi$
or $\delta$. We refer to answering the first question as a
\emph{basic} contrastive explanation, whereas answering the second
question will be referred to as a \emph{targeted} contrastive
explanation, and written as $\tn{CXp}_{\delta}$.
The duality result between AXps and CXps in~\autoref{th:dual}
applies \emph{only} to basic contrastive explanations.
Nevertheless, the algorithms for MCS extraction for computing a basic
CXp can also be adapted to computing targeted CXps, as follows.
We want a pick of feature values such that the prediction is
$\delta$. We start by letting all features to take any value, and such
that the resulting prediction is $\delta$. We then iteratively attempt
to fix feature values to those in the given instance, while the
prediction remains $\delta$. This way, the set of literals that change
value are a subset-minimal set of feature-value pairs that is
sufficient for predicting $\delta$.
Finally, there are crucial differences between the duality result
established in this section, which targets local explanations, and a
recent result~\cite{inms-nips19}, which targets \emph{global}
explanations.
Earlier work established a relation between prime implicants and
implicates as a way to relate global abductive explanations and
so-called counterexamples. In contrast, we delved into the
fundamentals of reasoning about inconsistency, concretely the duality
between MCSes and MUSes, and established a relation between
model-based \emph{local} AXps and CXps. 

%% file: algs.tex


%% file: res.tex
\section{Experimental Evaluation} \label{sec:res}
This section details the experimental evaluation to  assess
the practical feasibility and efficiency of the enumeration of
abductive and contrastive explanations for a few real-world datasets,
 studied in the context of explainability and algorithmic
fairness. To perform the evaluation, we adapt powerful algorithms for enumeration
MCSes or  MCSes and MUSes to find all abductive and
contrastive
explanations~\cite{stuckey-padl05,liffiton-jar08,lagniez-ijcai18,bendik-atva18}~\footnote{The prototype and the experimental setup are
available at \url{https://github.com/alexeyignatiev/xdual}.}. 
%
\autoref{alg:enum1}
and \autoref{alg:enum2} in \autoref{app:sec:algs} show our adaptations of MCS (resp.\ MCS and MUS) enumeration algorithms to the enumeration of CXps (resp.\ AXps and CXps).

\begin{figure}[h]
\centering
\hspace{-1.5em}
\begin{subfigure}{.16\textwidth}
  \includegraphics[width=1.4\linewidth]{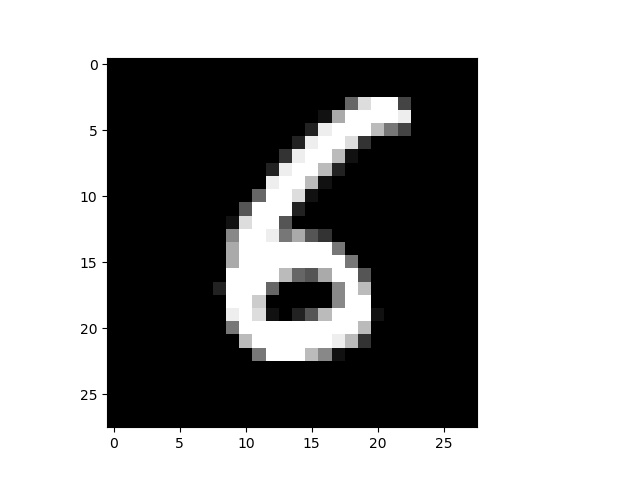}
  \caption{Real 6} \label{exp1:real}
\end{subfigure}
\begin{subfigure}{.16\textwidth}
   \includegraphics[width=1.4\linewidth]{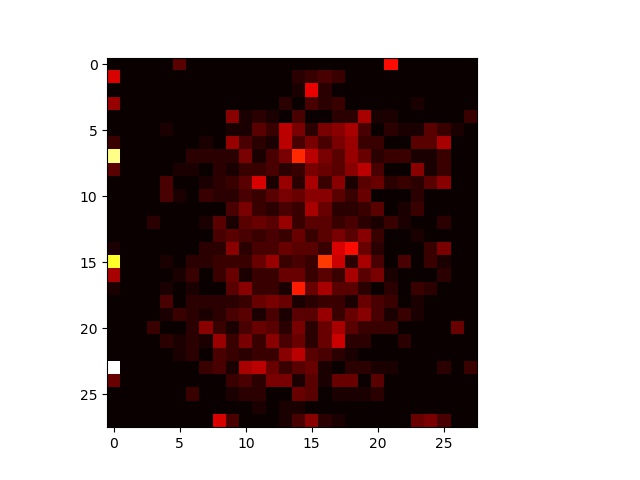}
  \caption{XGBoost} \label{exp1:real:xgboost}
\end{subfigure}
\begin{subfigure}{.16\textwidth}
  \includegraphics[width=1.4\linewidth]{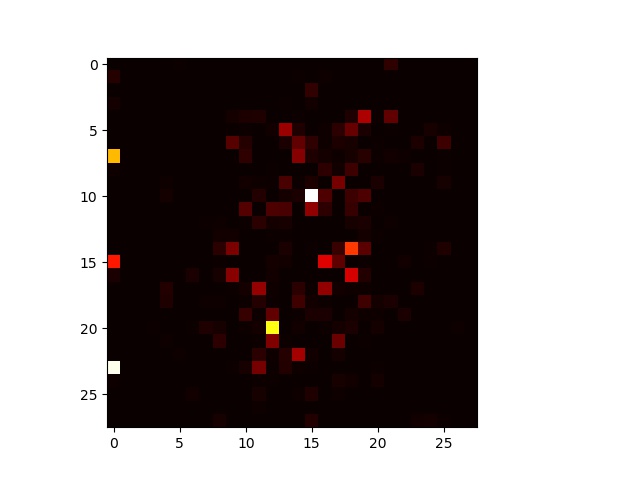}
  \caption{SHAP} \label{exp1:real:shap}
\end{subfigure}
\begin{subfigure}{.16\textwidth}
  \includegraphics[width=1.4\linewidth]{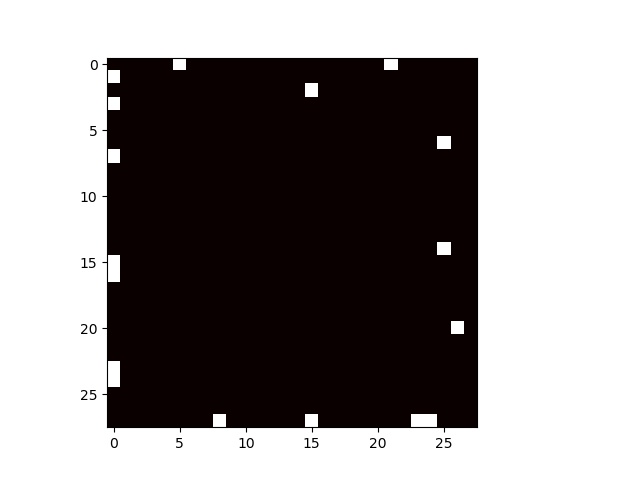}
  \caption{CXp$^\text{1}$}\label{exp1:real:mcses_1}
\end{subfigure}
\begin{subfigure}{.16\textwidth}
  \includegraphics[width=1.4\linewidth]{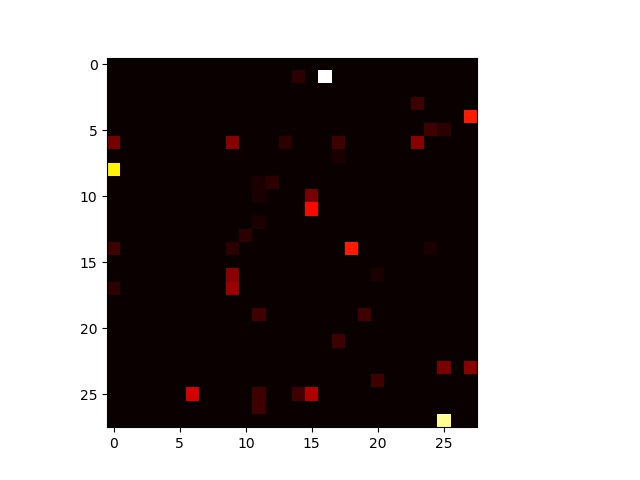}
  \caption{CXp$^\text{2}$}\label{exp1:real:mcses_2}
\end{subfigure}
\begin{subfigure}{.16\textwidth}
  \includegraphics[width=1.4\linewidth]{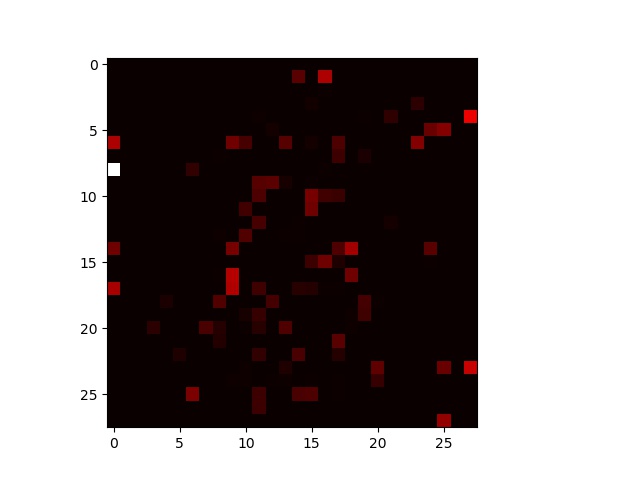}
  \caption{CXp$^\text{1--3}$}\label{exp1:real:mcses_all}
 \end{subfigure}\\
\hspace{-1.5em}
\begin{subfigure}{.16\textwidth}
  \includegraphics[width=1.4\linewidth]{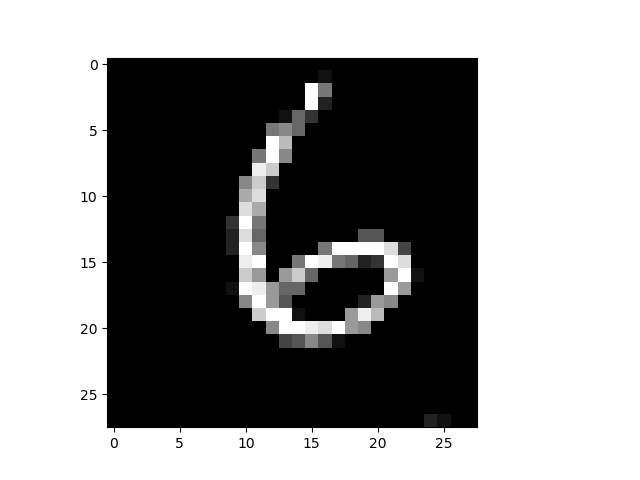}
  \caption{Fake 6}  \label{exp1:fake}
\end{subfigure}
\begin{subfigure}{.16\textwidth}
   \includegraphics[width=1.4\linewidth]{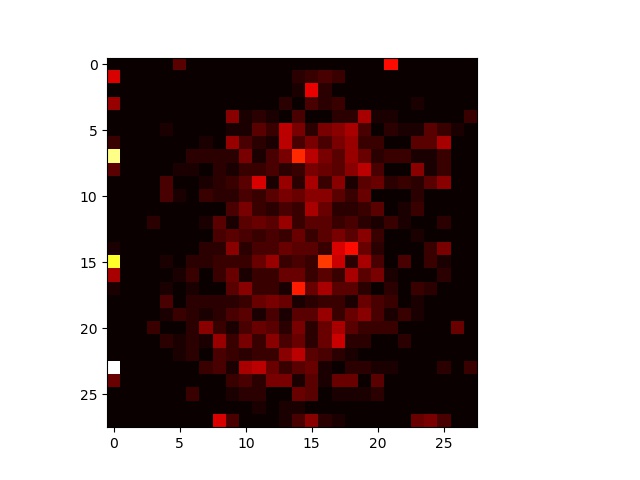}
  \caption{XGBoost}\label{exp1:fake:xgboost}
\end{subfigure}
\begin{subfigure}{.16\textwidth}
  \includegraphics[width=1.4\linewidth]{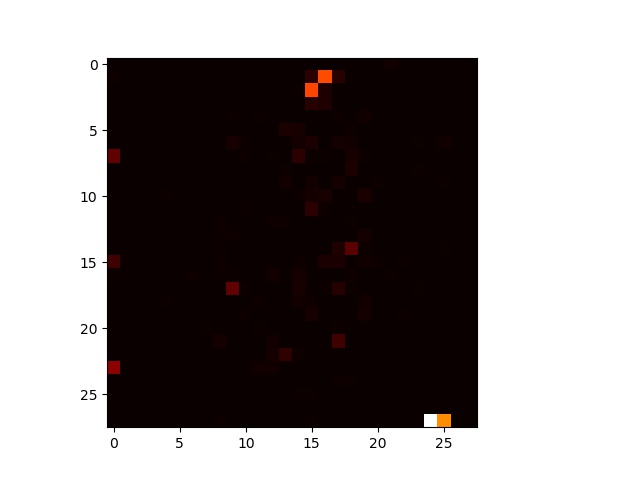}
 \caption{SHAP} \label{exp1:fake:shap}
\end{subfigure}
\begin{subfigure}{.16\textwidth}
  \includegraphics[width=1.4\linewidth]{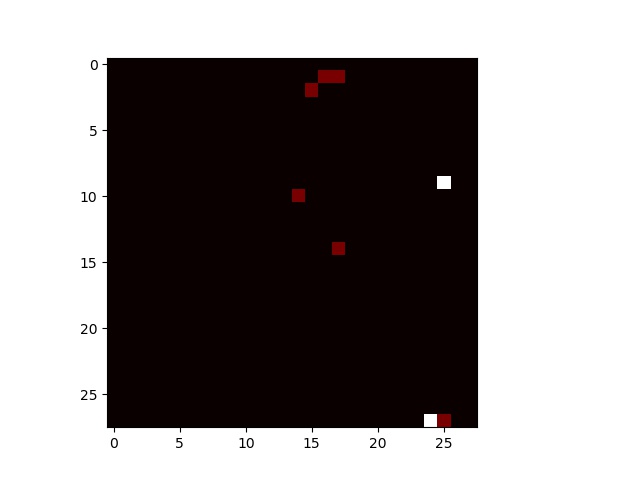}
  \caption{CXp$^\text{3}$}\label{exp1:fake:mcses_1}
\end{subfigure}
\begin{subfigure}{.16\textwidth}
  \includegraphics[width=1.4\linewidth]{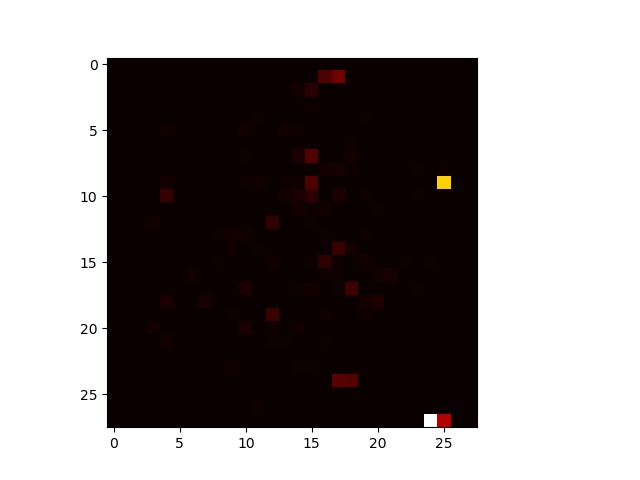}
  \caption{CXp$^\text{4}$}\label{exp1:fake:mcses_2}
 \end{subfigure}
\begin{subfigure}{.16\textwidth}
  \includegraphics[width=1.4\linewidth]{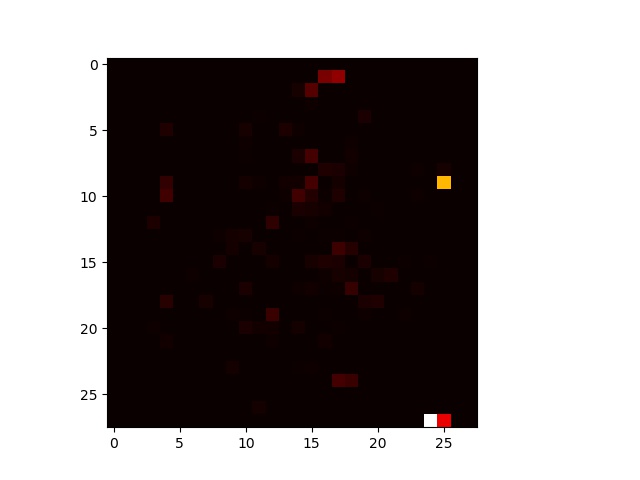}
  \caption{CXp$^\text{3--5}$}\label{exp1:fake:mcses_all}
\end{subfigure}
\vspace{-0.5\baselineskip}
\caption{The `real vs fake' images.
The first row shows results for the real image 6;
the second --  results for the fake image 6.
The first column shows examples of inputs; the second  --  heatmaps of XGBoost's important features; the third -- heatmaps of SHAP's explanation. Last three columns show heatmaps of  CXp of different cardinality.
The brighter pixels are more influential features.} \label{fig:gan}
\end{figure}
\vspace*{-7pt}
\paragraph{Enumeration of CXps.}
These experiments demonstrate a novel, unexpected practical use case of  CXps enumeration algorithms.
In particular, we show that our method gives a \emph{new fine-grained view} on both global and local standard explanations extracted from ML models. The goal of these experiments is to \emph{gain better understanding of existing explainers} rather than generate all CXps for a given input. We conduct two sets of experiments.
The first experiment, called ``real vs fake'', distinguishes real from fake images. A dataset contains two classes of images: (a) original MNIST digits and (b) fake MNIST digits produced by a standard DCGAN model~\cite{RadfordMC15} (see \autoref{exp1:real} and \autoref{exp1:fake} for typical examples). The second experiment, called ``3 vs 5 digits'', uses a dataset that contains digits ``3'' and ``5'' from the standard MNIST dataset (discussed in \autoref{app:sec:extra:exp:one}).
Next, we discuss the results of the ``real vs fake'' experiment in details (\autoref{fig:gan}).
For ``real vs fake'', we train an XGBoost model~\cite{guestrin-kdd16b}  with 100 trees of depth 6  (accuracy 0.85/0.80 on train/test sets). We quantized images so that each pixel takes a value  between 0 and 15, image pixels are categorical features  in the model. 

\textit{Brief overview of the SHAP explainer.}
Given a classifier $f$ and an explainer model $g$, SHAP aims to train $g$ be similar to $f$ in the neighborhood of some given point $x$.
The objective function for SHAP is designed so that: (1) $g$ approximates the behavior of the black box $f$ accurately within the vicinity of $x$, and (2) $g$ achieves lower complexity and is  interpretable:
$\xi(x) = \argmin_{g\in G} \,\, L(\pi_x,g,f) + \Omega(g)$,
where the loss function $L$ is defined  to minimize the distance between $f$ and $g$ in the neighborhood of $x$ using a weight function $\pi_x$ and $\Omega(g)$ quantifies the complexity of $g$; $\Omega(g)$ and $\pi_x$ are defined based on game-theoretic notions~\cite{lundberg-nips17}.

\vspace{-2pt}
\textit{Global and local explainers.}
We start by discussing our results on a few samples
(\autoref{exp1:real} and \autoref{exp1:fake}). First, we extract important features provided by XGBoost.
As these features are \emph{global} for the model, they are the same for all inputs (\autoref{exp1:real:xgboost} and \autoref{exp1:fake:xgboost} are identical
for real and fake images). \autoref{exp1:real:xgboost}  shows that these important features are \emph{no very informative} for this dataset as these pixels
form a blob of pixels that cover an image.
Then we compute an image-specific explanation using the standard explainer SHAP (see  \autoref{exp1:real:shap} for the real image and \autoref{exp1:fake:shap} for the fake image).

SHAP explanations are more focused on specific parts of images compared to XGBoost. However, it is still not easy to  gain insights about which areas of an image are more important as pixels all over the image participate in the explanations of SHAP and XGBoost.
For example, both XGBoost and SHAP distinguish some edge and middle pixels as key pixels (the bright pixels are  more important) but it is not clear why these are important pixels.

\vspace{-2pt}
\textit{CXps enumeration approach.}
We recall that our goal is to investigate whether there is a connection between  the important pixels that SHAP/XGBoost finds and CXps for a given image. The most surprising result is that, indeed, a connection exists and, for example, it reveals that the edge pixels of an image, highlighted by both SHAP and XGBoost as important pixels,  are, in fact, CXps of small cardinalities.
Given all CXps of size $k$, we plot a heatmap of occurrences of each pixel in these CXps of size $k$.
Let us focus on the first row with the real 6.
Consider the heatmap CXp$^\text{1}$ at \autoref{exp1:real:mcses_1} that shows all CXps of size one for the real 6. It shows that most of important pixels of XGBoost and SHAP are actually CXps of size one. This means that \emph{it is sufficient to change a single pixel value to some other value to obtain a different prediction}. 
Note that these results reveal an interesting observation. DCGAN generates images with a few gray edges pixels (see \autoref{fig:extra} in Appendix. Indeed, some of them have several edge pixels in gray.) This `defect' does not happen often for real MNIST images.  Therefore, the classifier `hooks' on this issue to classify an image as fake.
Now, consider the heatmap CXp$^\text{2}$ at  \autoref{exp1:real:mcses_2} of CXps of size two.
It overlaps a lot with SHAP important pixels in the middle of the image explaining \emph{why} these are important. Only a \emph{pair} of these pixels can be changed to get a different prediction.

\textit{A correlation between CXps and SHAP's important  features.}
To qualitatively measure our observations on correlation between key
features of CXps and SHAP, we conducted the same experiment as above
on 100 random images and measured the correlation between CXps and
SHAP features.
First, we compute a set $T$ of pixels that is the union of the first
(top) 100 smallest size CXps.
On average, we have 60 pixels in $T$.
Note that the average 60 pixels represent a small fraction (7\%) of the total
number of pixels.
Then we find a set $S$ of $|T|$ SHAP pixels with highest absolute
weights.
Finally, we compute $corr = {|S \cap T|}/{|S|}$ as the correlation
measure.
Note that $corr=\text{0.4}$ on average, i.e.\ our method hits 40\% of best
SHAP features.
As the chances of two tools independently hitting the same pixel (out
of 784) are quite low, the fact that 40\% of $|S|$ are picked
indicates a significant correlation.

%
\vspace*{-7pt} \paragraph{Enumeration of CXps and AXps.}
\input{./tabs/restab1}
%
Here, we aim at testing the \emph{scalability} of explanation
enumeration and consider the six well-known and publicly available
datasets.
Three of them were previously studied in~\cite{guestrin-aaai18} in the
context of heuristic explanation approaches, namely,
Anchor~\cite{guestrin-aaai18} and LIME~\cite{guestrin-kdd16},
including \emph{Adult}, \emph{Lending}, and \emph{Recidivism}.
\autoref{app:sec:extra:cp_ap} provides a detailed explanation of
datasets and our implementation.
A prototype implementing is an adaptation of~\cite{liffiton-jar08}
abductive or (2)~all contrastive explanations was created.
In the experiment, the prototype implementation is instructed to
enumerate all abductive explanations.
%
The prototype is able to deal with tree ensemble models trained with
XGBoost~\cite{guestrin-kdd16b}.
%
%
%
%
%
Given a dataset, we trained an XGBoost model containing 50 trees per
class, each tree having depth 3.
(Further increasing the number of trees per class and also increasing
the maximum depth of a tree did not result in a significant increase
of the models’ accuracy on the training and test sets for the
considered datasets.)
All abductive explanations for every instance of each of the six
datasets were exhaustively enumerated using the duality-based approach
(\autoref{alg:enum2} in \autoref{app:sec:algs}). This resulted in the
computation of all contrastive explanations as well).

\textit{Evaluation results.}
%
%
\autoref{tab:res} shows the results.
There are several points to make.
First, although it seems computationally expensive to enumerate all
explanations for a data instance, it can still be achieved effectively
for the medium-sized models trained for all the considered datasets.
This may on average require from a few dozen to several hundred of
oracle calls per data instance (in some cases, the number of calls
gets up to a few thousand).
Also observe that enumerating all explanations for an instance takes
from a fraction of a second to a couple of seconds on average. These results demonstrate that our approach is practical.

Second, the total number of AXps 
is typically lower
than the total number of their contrastive counterparts.
The same holds for the average numbers of abductive and contrastive
explanations per data instance.
Third and finally, AXps
for the studied datasets
tend to be larger than contrastive explanations.
The latter observations imply that contrastive explanations may be
preferred from a user's perspective, as the smaller the explanation is
the easier it is to interpret for a human decision maker.
(Furthermore, although it is not shown in \autoref{tab:res}, we
noticed that in many cases contrastive explanations tend to be of size
1, which makes them ideal to reason about the behaviour of an ML
model.)
On the other hand, exhaustive enumeration of contrastive explanations
can be more time consuming because of their large number.
\vspace*{-6pt}
\paragraph{Summary of results.}
We show that  CXps enumeration gives us an insightful understanding of a classifier's behaviour. First, even in cases when we cannot enumerate all of CXps to compute AXps by duality,
we can still draw some conclusions, e.g.  CXps of size one are exactly features that occur in all AXps.
Next, we clearly demonstrate the
feasibility of the duality-based exhaustive enumeration of both AXps and CXps for a given data instance using a more powerful algorithm that performs enumeration of AXps and CXps.

%% file: tabs/restab1.tex
\begin{table*}[t]
  \caption{Results of the computational experiment on enumeration of AXps and CXps.}
  \scriptsize
  \begin{adjustbox}{center}
  \sisetup{math-rm=\textrm}
  \setlength{\tabcolsep}{1.6em}
  \begin{tabular}{rS[table-format=2.1]S[table-format=2.1]S[table-format=2.1]S[table-format=2.1]S[table-format=2.1]S[table-format=2.1]}
    \toprule
    & \multicolumn{6}{c}{\textbf{Dataset}} \\ \cmidrule{2-7}
    & \textbf{Adult} & \textbf{Lending} & \textbf{Recidivism} & \textbf{Compas} & \textbf{German} & \textbf{Spambase} \\ \midrule
    \textbf{\# of instances} & 5579. & 4414. & 3696. & 778. & 1000. & 2344.0 \\ \midrule
    \textbf{total time (sec.)} & 7666.9 & 443.8 & 3688.0 & 78.4 & 16943.2 & 6859.2 \\
    \textbf{minimal time (sec.)} & 0.1 & 0.0 & 0.1 & 0.0 & 0.2 & 0.1 \\
    \textbf{average time (sec.)} & 1.4 & 0.1 & 1.0 & 0.1 & 16.9 & 2.9 \\
    \textbf{maximal time (sec.)} & 13.1 & 0.8 & 8.9 & 0.5 & 193.0 & 23.1 \\ \midrule

    \textbf{total oracle calls} & 492990. & 69653. & 581716. & 21227. & 748164. & 176354. \\
    \textbf{minimal oracle calls} & 14. & 11. & 17. & 13. & 23. & 12. \\
    \textbf{average oracle calls} & 88.4 & 15.8 & 157.4 & 27.3 & 748.2 & 75.2 \\
    \textbf{maximal oracle calls} & 581. & 73. & 1426. & 134. & 7829. & 353. \\ \midrule

    \textbf{total \# of AXps} & 52137. & 8105. & 60688. & 1931.0 & 59222. & 18876. \\
    \textbf{average \# of AXps} & 9.4 & 1.8 & 16.4 & 2.5 & 59.2 & 8.1 \\
    \textbf{average AXp size} & 5.3 & 1.9 & 6.4 & 3.8 & 7.5 & 4.6 \\ \midrule

    \textbf{total \# of CXps} & 66219. & 8663. & 77784. & 3558.0 & 66781. & 24774. \\
    \textbf{average \# of CXps} & 11.9 & 2.0 & 21.1 & 4.6 & 66.8 & 10.6 \\
    \textbf{average CXp size} & 2.4 & 1.4 & 2.6 & 1.5 & 3.6 & 2.3 \\ \bottomrule
  \end{tabular}
  \end{adjustbox} \label{tab:res}
\end{table*}

%% file: conc.tex
\section{Conclusions} \label{sec:conc}
This paper studies local model-based abductive and contrastive
explanations. Abductive explanations answer
\axpdef
questions, whereas contrastive explanations answer
\cxpdef
questions.
Moreover, the paper relates explanations with the analysis of
inconsistent theories, and shows that abductive explanations
correspond to minimal unsatisfiable subsets, whereas contrastive
explanations can be related with  minimal correction subsets.
As a consequence of this result, the paper exploits a well-known
minimal hitting set relationship between MUSes and
MCSes~\cite{reiter-aij87,lozinskii-jetai03} to reveal the same
relationship between abductive and contrastive explanations.
In addition, the paper exploits known results on the analysis of
inconsistent theories, to devise algorithms for extracting and
enumerating abductive and contrastive explanations.

%% file: appendix.tex
\newpage
\appendix

\section{Taxonomy}
\label{app:taxonomy}
The taxonomy of explanations used in the paper is summarized in~\autoref{tab:taxo}.
\input{taxonomy}

\section{Extracting \& Enumerating Explanations} \label{app:sec:algs}

The results of~\autoref{sec:dual} enable exploiting past work on
extracting and enumerating MCSes and MUSes to the setting of
contrastive and abductive explanations, respectively.
Perhaps surprisingly, there is a stark difference between algorithms
for extraction and enumeration of contrastive explanations and
abductive explanations.
Due to the association with MCSes, one contrastive explanation can be
computed with a logarithmic number of calls to a decision
procedure~\cite{liffiton-jar08}.
Moreover, there exist algorithms for the direct enumeration of
contrastive explanations~\cite{liffiton-jar08}.
In contrast, abductive explanations are associated with MUSes. As a
result, any known algorithm for extraction of one abductive
explanation requires at best a linear number of calls to a decision
procedure~\cite{junker04}, in the worst-case. 
Moreover, there is no known algorithm for the direct enumeration of
abductive explanations, and so enumeration can be achieved only
through the enumeration of contrastive
explanations~\cite{liffiton-jar08,lpmms-cj16,Felfernig2012}.

We adapt state-of-the-art algorithms for the enumeration MUSes and
MCSes to find all the abductive and contrastive explanations. Note
that as in the case of enumeration of MCSes and MUSes, the enumeration
of CXps is comparatively easier than the enumeration of AXps.
\autoref{alg:enum1} shows our adaptation of MCS enumeration algorithm
to the enumeration of CXps~\cite{liffiton-jar08}. Other
alternatives~\cite{lagniez-ijcai18} could be considered instead.
%
\autoref{alg:enum1} finds a CXp, blocks it and finds the next one
until no more exists. To extract a single CXp, we can use standard
algorithm, e.g.\ \cite{stuckey-padl05}.  In principle, 
enumeration of AXps can be achieved by computing all CXps and then
computing all the minimal hitting sets of all CXps, as proposed in
the propositional setting~\cite{liffiton-jar08}. However, there are
more efficient alternatives that we can adapt
here~\cite{stuckey-padl05,lpmms-cj16,narodytska-ijcai18,bendik-atva18},
\autoref{alg:enum2} adapts~\cite{lpmms-cj16} to
the case of computing both AXps and CXps. The algorithm simultaneously searches for AXps and CXps and is based on the hitting set duality.

\begin{algorithm}[H] 
\small
  \input{algs/enum-cxp}
  \caption{Enumeration of CXps} \label{alg:enum1}
\end{algorithm}

\begin{algorithm}[H] 
    \small
  \input{algs/enum-axp-cxp}
  \caption{Enumeration of AXps (and CXps)} \label{alg:enum2}
\end{algorithm}

\section{Additional experimental results}
\subsection{Enumeration of CXps}\label{app:sec:extra:exp:one}

\begin{figure}[h]
\centering
\begin{subfigure}{.16\textwidth}
  \includegraphics[width=1.2\linewidth]{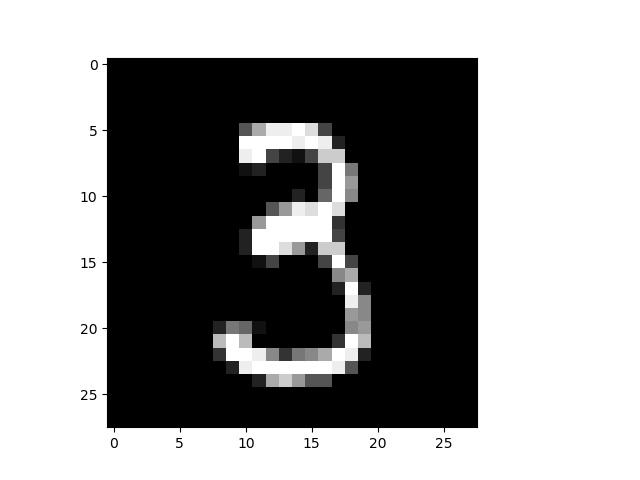}
  \caption{Digit 3}\label{exp2:digit3}
\end{subfigure}
\begin{subfigure}{.16\textwidth}
   \includegraphics[width=1.2\linewidth]{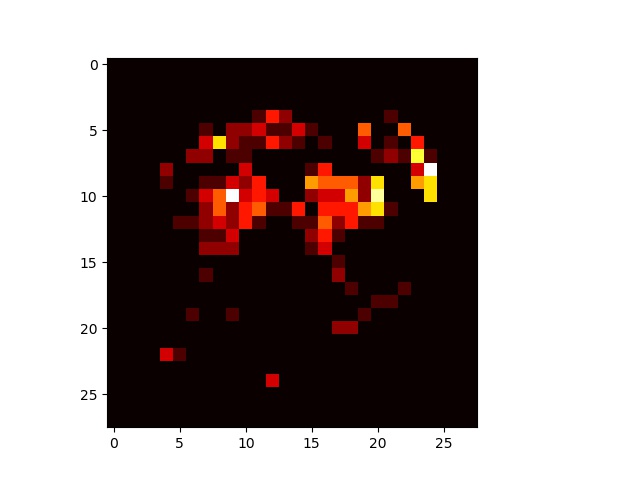}
  \caption{XGBoost}\label{exp2:3:xgboost}
\end{subfigure}
\begin{subfigure}{.16\textwidth}
  \includegraphics[width=1.2\linewidth]{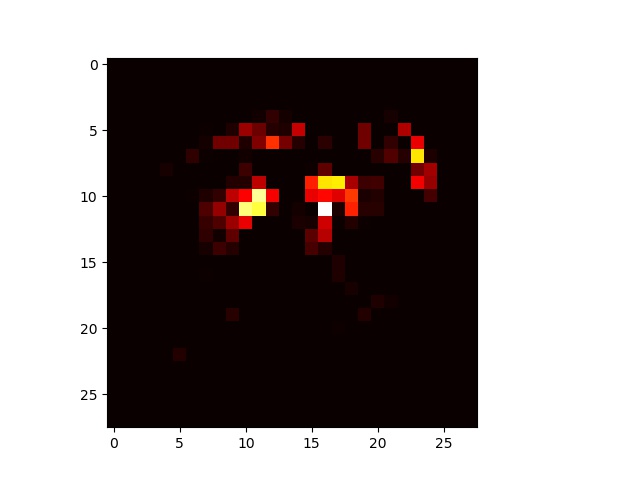}
  \caption{SHAP}\label{exp2:3:shap}
\end{subfigure}
\begin{subfigure}{.16\textwidth}
  \includegraphics[width=1.2\linewidth]{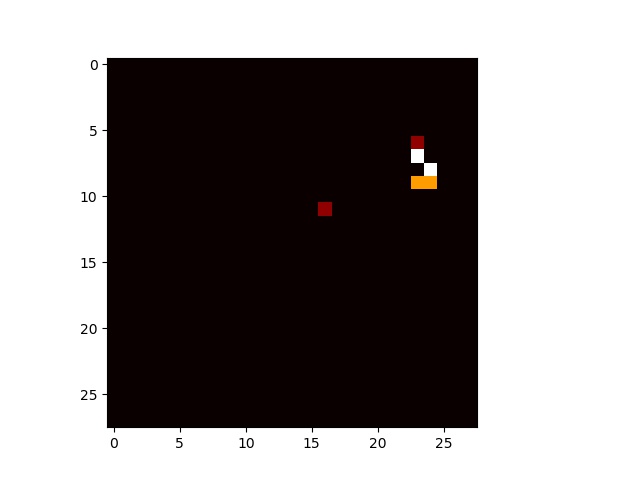}
  \caption{CXp$^\text{3}$}\label{exp2:3:mcses_1}
\end{subfigure}
\begin{subfigure}{.16\textwidth}
  \includegraphics[width=1.2\linewidth]{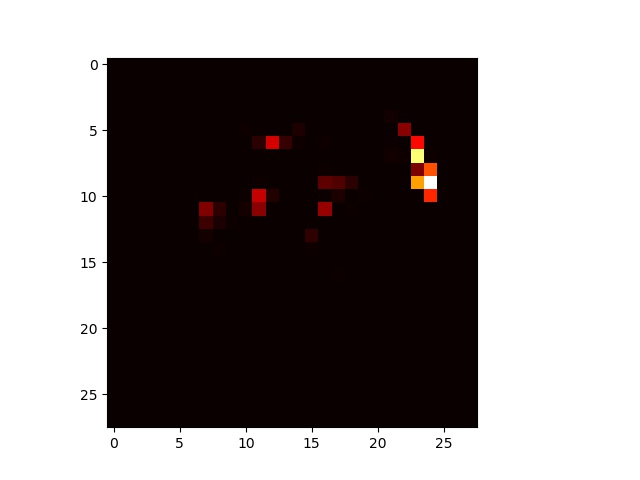}
  \caption{CXp$^\text{4}$}\label{exp2:3:mcses_2}
\end{subfigure}
\begin{subfigure}{.16\textwidth}
  \includegraphics[width=1.2\linewidth]{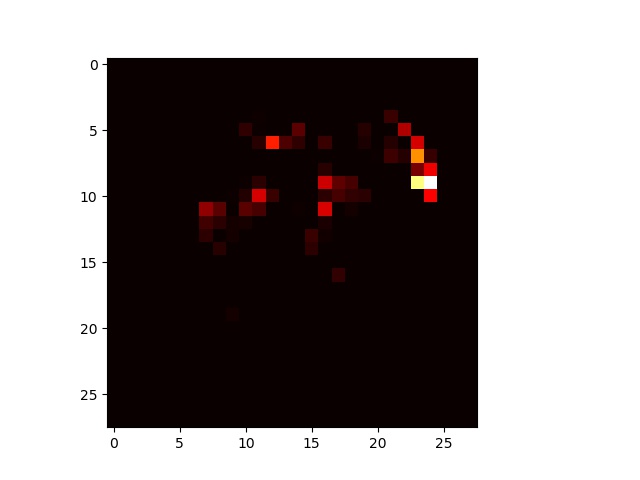}
  \caption{CXp$^\text{3--5}$}\label{exp2:3:mcses_all}
\end{subfigure}\\
\begin{subfigure}{.16\textwidth}
  \includegraphics[width=1.2\linewidth]{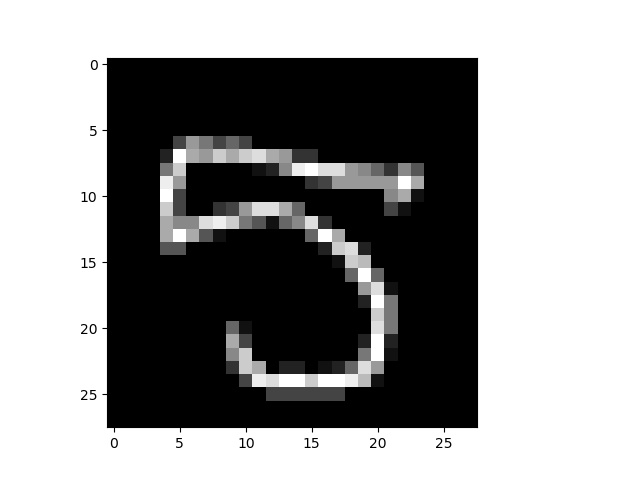}
  \caption{Digit 5}\label{exp2:digit5}
\end{subfigure}
\begin{subfigure}{.16\textwidth}
   \includegraphics[width=1.2\linewidth]{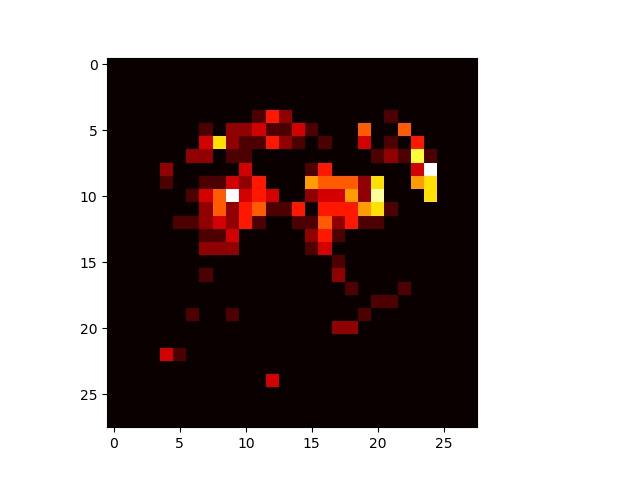}
  \caption{XGBoost}\label{exp2:5:xgboost}
\end{subfigure}
\begin{subfigure}{.16\textwidth}
  \includegraphics[width=1.2\linewidth]{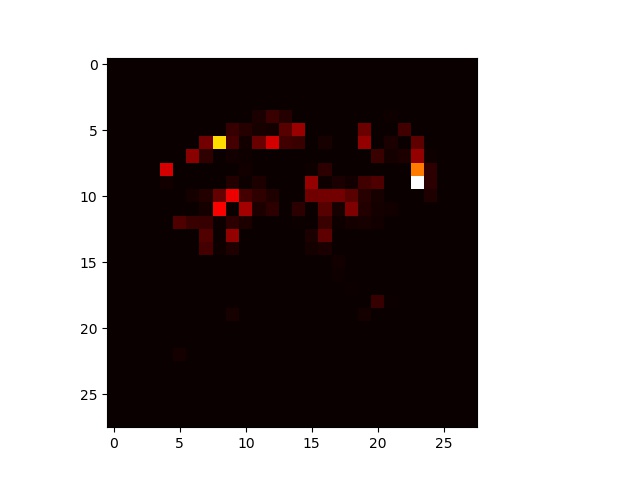}
  \caption{SHAP}\label{exp2:5:shap}
\end{subfigure}
\begin{subfigure}{.16\textwidth}
  \includegraphics[width=1.2\linewidth]{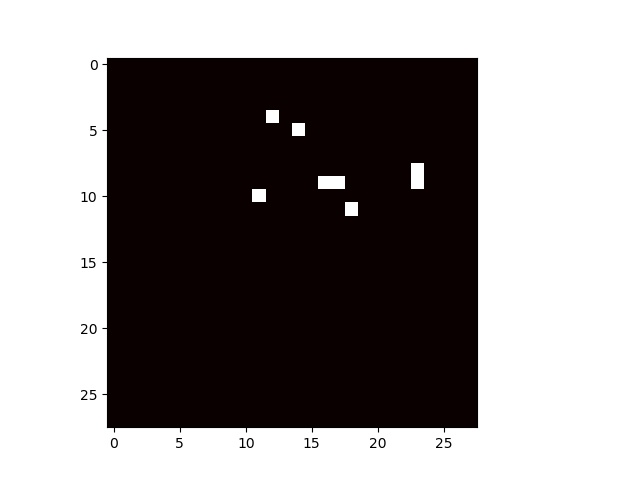}
    \caption{CXp$^\text{1}$}\label{exp2:5:mcses_1}
\end{subfigure}
\begin{subfigure}{.16\textwidth}
  \includegraphics[width=1.2\linewidth]{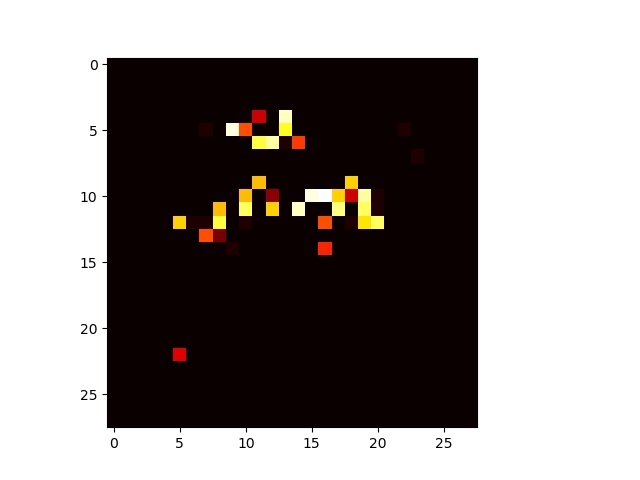}
  \caption{CXp$^\text{2}$}\label{exp2:5:mcses_2}
\end{subfigure}
\begin{subfigure}{.16\textwidth}
  \includegraphics[width=1.2\linewidth]{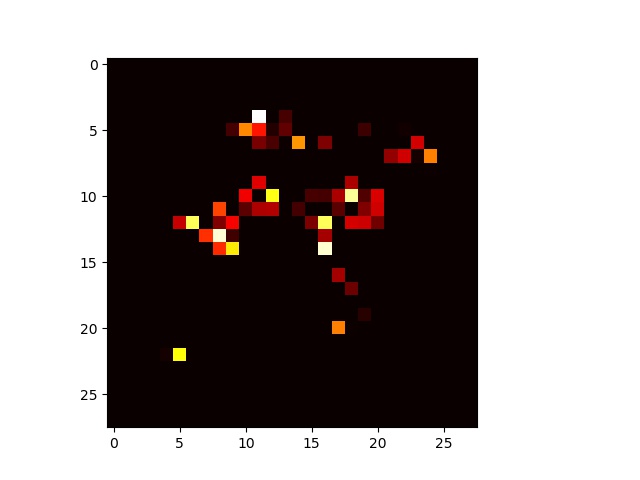}
    \caption{CXp$^\text{1--6}$}\label{exp2:5:mcses_all}
\end{subfigure}
\caption{Results of the 3 vs 5 digits experiments.
The first row shows results for the image 3.
The second row shows results for the image 5.
The first column shows examples of inputs; the second column shows  heatmaps of XGBoost's global important features; the third column shows heatmaps of SHAP's important features. Last three columns show heatmaps of CXp of different cardinality.
}\label{fig:digits}
\end{figure}

\paragraph{Setup.}
To perform enumeration of contrastive explanations in our first experiment, we use a constraint programming solver, ORtools~\cite{ortools}~\footnote{The prototype and the experimental setup are
available at \url{https://github.com/alexeyignatiev/xdual}.}. To encode the enumeration problem with ORtools we converted scores of XGBoost models into integers keeping 5 digits precision. We enumerate
contrastive explanations in the increasing order by their cardinality. This can be done by a simple modification of \autoref{alg:enum1} forcing it to return CXps in this order. So, we first obtain all minimum size
contrastive explanations, and so on.
\paragraph{Second experiment.}
Consider our second the ``3 vs 5 digits'' experiment. We use a dataset that contains  digits ``3'' (class 0)  and ``5'' (class 1) from the standard MNIST (see \autoref{exp2:digit3} and \autoref{exp2:digit5} for representative samples).
XGboost model has 50 trees of depth 3 with accuracy 0.98 (0.97) on train/test sets.
We quantized images so that each pixel takes a value  between 0 and 15.    As before, each pixel corresponds to a feature. So, we have 784 features in our XGBoost model.

\begin{figure}[h]
\centering
\begin{subfigure}{.2\textwidth}
  \includegraphics[width=1.2\linewidth]{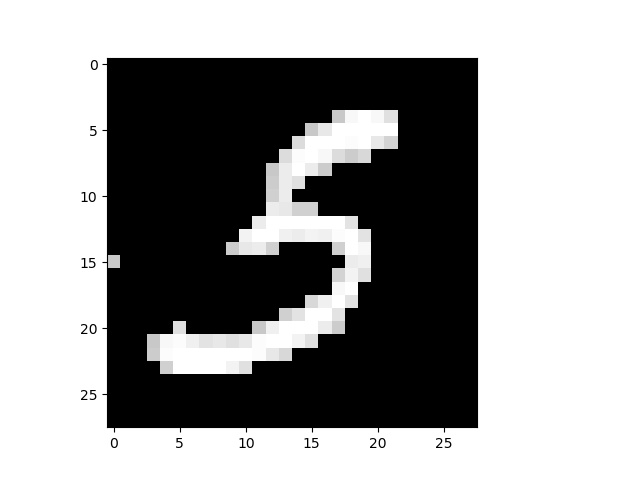}
\end{subfigure}
\begin{subfigure}{.2\textwidth}
   \includegraphics[width=1.2\linewidth]{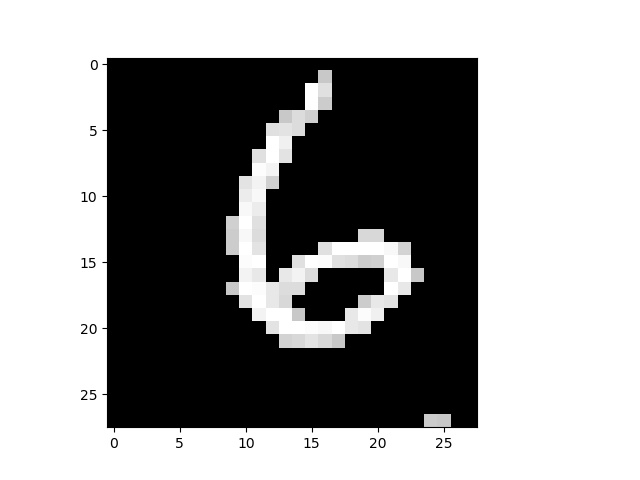}
\end{subfigure}
\begin{subfigure}{.2\textwidth}
  \includegraphics[width=1.2\linewidth]{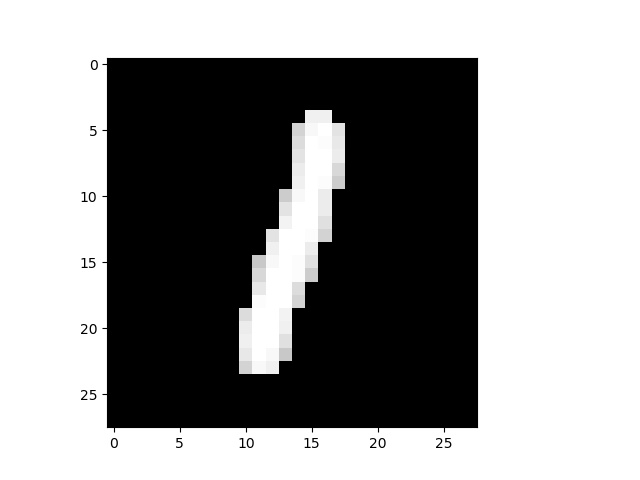}
\end{subfigure}
\begin{subfigure}{.2\textwidth}
  \includegraphics[width=1.2\linewidth]{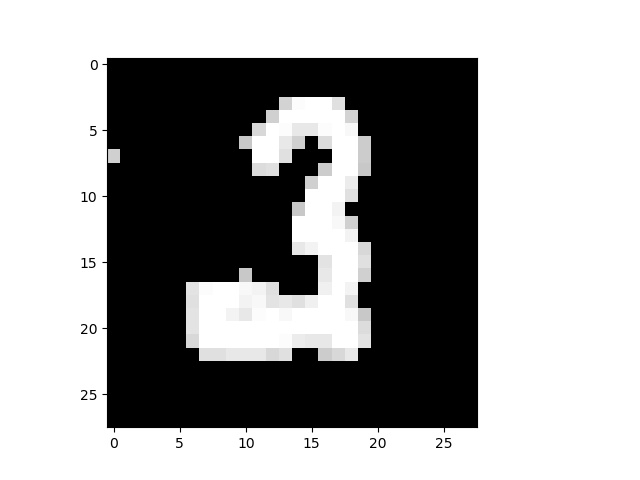}
\end{subfigure}
\caption{Additional fake images. We reduced values of zero-valued pixels to highlight gray pixels on the edges for some fake images.} \label{fig:extra}
\end{figure}

\paragraph{Global and local explainers.}
We start by discussing our results on few random samples
(\autoref{exp2:digit3} and \autoref{exp2:digit5}).
First, we obtain the important features from XGBoost.
As these features are \emph{global} for the model so they are the same for all inputs (\autoref{exp2:3:xgboost} and \autoref{exp2:5:xgboost} are identical
for 3  and 5  images). \autoref{exp1:real:xgboost}  shows that these important features. The important pixels highlight that the top parts of images are important, which is a plausible high-level explanation of the classifier behavior. Digits 3 and 5 are mostly differ in the top part of the image.  However, some pixels are way more important than other and it is hard to understand why.

Next, we compute an image-specific explanation using the standard explainer SHAP ( see  \autoref{exp2:3:shap} for the digit  3 and \autoref{exp2:3:shap} for the  digit 5). While SHAP explanations mimic  XGBoost important features, they do provide additional insights for the user.
Note that both XGBoots and SHAP mark a ``belt'' of pixels in the upper middle part that as important (bright pixels is the most important pixels).

\paragraph{CXps enumeration approach.}
We run our enumeration algorithm to produce CXps of increasing cardinality. For each image, we enumerate first 2000 CXps.
Given all CXps of size $k$, we plot a heatmap of occurrences of each pixel in these CXps of size $k$.
Let us focus on the second row with the digit 5.
For example, CXp$^\text{2}$ (\autoref{exp2:5:mcses_2}) shows the heatmap of CXps of size two for the digit 5.
As we mentioned above, both XGBoost and SHAP hint that the `belt' of  important pixels in the middle.
Again, our method can explain \emph{why} this is the case.
Consider the heatmap CXp$^\text{1}$ at \autoref{exp2:5:mcses_1}. This picture shows all CXps of size one for the digit 5. It reveals that most of important pixels of XGBoost and SHAP are actually CXps of size one. We reiterate that  it is sufficient to change a \emph{single} pixel value to some other value to obtain a different prediction.
Now, consider the heatmap CXp$^\text{1--6}$ at  \autoref{exp2:5:mcses_all}. This figure shows 2000 CXps (from size 1 to size 6). It overlaps a lot with SHAP important pixels in the middle of the image. So, these pixels occur in many small size CXps and changing their values leads to misclassification.

\paragraph{Correlation between CXps and SHAP features.}
To qualitatively measure our observations on correlation between key
features of CXps and SHAP, we conducted the same experiment as above
on 100 random images and measured the correlation between CXps and
SHAP features.
First, we compute a set $T$ of pixels that is the union of the first
(top) 100 smallest size CXps.
On average, we have 38 pixels in $T$.
Note that the average 38 pixels represent a small fraction (5\%) of the total
number of pixels.
Then we find a set $S$ of $|S|$ SHAP pixels with highest absolute
weights.
Finally, we compute $corr = {|S \cap T|}/{|S|}$ as the correlation
measure.
Note that $corr=\text{0.6}$ on average, i.e.\ our method hits 60\% of best
SHAP features.
As the chances of two tools independently hitting the same pixel (out
of 784) are quite low, the fact that 60\% of $|T|$ are picked
indicates a significant correlation.

\subsection{Enumeration of CXps and AXps}\label{app:sec:extra:cp_ap}

\paragraph{Datasets.}
%
The results are obtained on the six well-known and publicly available
datasets.
Three of them were previously studied in~\cite{guestrin-aaai18} in the
context of heuristic explanation approaches, namely,
Anchor~\cite{guestrin-aaai18} and LIME~\cite{guestrin-kdd16},
including \emph{Adult}, \emph{Lending}, and \emph{Recidivism}.
These datasets were processed the same way as
in~\cite{guestrin-aaai18}.
The \emph{Adult} dataset~\cite{kohavi-kdd96} is originally taken from
the Census bureau and targets predicting whether or not a given adult
person earns more than \$50K a year depending on various attributes,
e.g.\ education, hours of work, etc.
The \emph{Lending} dataset aims at predicting whether or not a loan on
the Lending Club website will turn out bad.
The \emph{Recidivism} dataset was used to predict recidivism for
individuals released from North Carolina prisons in 1978 and
1980~\cite{schmidt-1988}.
Two more datasets were additionally considered including \emph{Compas}
and \emph{German} that were previously studied in the context of the
FairML and Algorithmic Fairness
projects~\cite{fairml17,fairness15,feldman-kdd15,friedler-fat19}, an
area in which the need for explanations is doubtless.
\emph{Compas} is a popular dataset, known~\cite{propublica16} for
exhibiting racial bias of the COMPAS algorithm used for scoring
criminal defendant's likelihood of reoffending.
The latter dataset is a German credit data (e.g.\
see~\cite{feldman-kdd15,friedler-fat19}), which given a list of
people's attributes classifies them as good or bad credit risks.
Finally,  we consider the \emph{Spambase} dataset from the UCI
repository~\cite{Dua2019}.
The main goal is to classify an email as spam or non-spam based on the
words that occur in this email.
Due to scalability constraints, we preprocessed the dataset to keep
ten words per email that were identified as the most influential words
by a random forest classifier.

\paragraph{Implementation and Setup.}
A prototype implementing \autoref{alg:enum2} targeting the enumeration
of either (1)~all abductive or (2)~all contrastive explanations was
created.
In the experiment, the prototype implementation is instructed to
enumerate all abductive explanations.
(Note that, as was also mentioned before, no matter what kind of
explanations \autoref{alg:enum2} aims for, all the dual explanations
are to be computed as a side effect of the hitting set duality.)
The prototype is able to deal with tree ensemble models trained with
XGBoost~\cite{guestrin-kdd16b}.
For that purpose, a simple encoding of tree ensembles into
satisfiability modulo theories (SMT) was developed.
Concretely, the target formulas are in the theory of linear arithmetic
over reals (RIA formulas).
(Note that encodings of a decision tree into logic are
known~\cite{bonfietti-cpaior15,lombardi-aij17,verwer-aij17}.
%
The final score summations used in tree ensembles can be encoded into RIA formulas.)

Due to the twofold nature of \autoref{alg:enum2}, it has to deal with
(1)~implicit hitting set enumeration and (2)~entailment queries with
SMT.
%
%
The former part is implemented using the award-winning maximum
satisfiability solver RC2~\cite{imms-jsat19} written on top of the
PySAT toolkit~\cite{imms-sat18}.
SMT solvers are accessed through the PySMT framework~\cite{gm-smt15},
which provides a unified interface to a variety of state-of-the-art
SMT solvers.
In the experiments, we use Z3~\cite{demoura-tacas08} as one of the
best performing SMT solvers.
The conducted experiment was performed in Debian Linux on an Intel
Xeon~E5-2630 2.60GHz processor with 64GByte of memory.


%% file: taxonomy.tex
\begin{table}[H]  
  \centering
  \caption{Taxonomy of ML model explanations used in the paper.}
  \label{tab:taxo}
\resizebox{\columnwidth}{!}{%
  \begin{tabular}{cc|p{6cm}|c|}
    \cline{3-4}
    &&\multicolumn{2}{c|}{ \textbf{Instance-}}\\
    \cline{3-4}
    & & \multicolumn{1}{c|}{\textit{dependent}} & \multicolumn{1}{c|}{\textit{independent}}  \\
    \cline{3-4}
    \hline
    \multicolumn{1}{|c|}{ \multirow{2}{*}{\rotatebox[origin=c]{90}{ \textbf{ML model-}
    }}} &
    \rotatebox[origin=c]{90}{  {agnostic} } &
    \pbox{10cm}{ Heuristic \textit{local} explanation  for $\pi$.\\
      Examples: SHAP, LIME,  Anchor, etc.}
    & \pbox[c]{10cm}{ {Heuristic}  \textit{global} explanation  for $\pi$.\\
      Examples: SHAP, LIME (e.g. submodular pick)}\\
    \cline{2-4}
    \multicolumn{1}{|c|}{} &\rotatebox[origin=c]{90}{  {based} }&
    \pbox[c]{10cm}{
      \vphantom{$c^C$} \enspace  Rigorous \textit{local} explanation  for $\pi$.\\
      \vphantom{$c^C$} \enspace   Examples: \\
      \begin{tabular}{c|c}
        \hline
        `Why $\pi$?' & \textcolor{black}{`Why not $\neg \pi$ ?'}\\
        \hline
         PI- (abductive)  & \textcolor{black}{ contrastive (CXps)} \\
          explanations (AXps)  & \textcolor{blue}{(our work)}
      \end{tabular}
    }
    & \pbox[c]{10cm}{ Rigorous  \textit{global} explanation for $\pi$.\\
      Examples: absolute/global AXps}\\
    \hline
  \end{tabular}
}
\end{table}

%% file: algs/enum-cxp.tex
%
%
\SetKwFunction{tfunc}{{\sc CXpEnum}} %
\Func \tfunc{$\fml{M}_{\pi}$,$\fml{C}$, $\pi$} \;

\Indp
\KwIn{
  $\fml{M}_{\pi}$: ML model,
  $\fml{C}$: Input cube,
  $\pi$: Prediction
}
\KwVars{
  $\fml{N}$ and $\fml{P}$ defined on the variables of $\fml{C}$
}
\TopBlankLine
{
  \lnlset{a3:01}{1}%
  $\fml{I}\gets\emptyset$ \tcp*[r]{Block CXps}
  \lnlset{a3:02}{2}%
  \While{\true}{
    \lnlset{a3:03}{3}%
    $\mu\gets\ExtractCXp(\fml{M}_{\pi},\fml{C},\pi,\fml{I})$\;
    
    \lnlset{a3:04}{4}%
    \lIf{$\mu=\emptyset$}{\Break} 
    
    \lnlset{a3:05}{5}%
    $\ReportCXp(\mu)$\;

    \lnlset{a3:06}{6}%
    $\fml{I}\gets\fml{I}\cup\NegLits(\mu)$ \;
  }
}
\Indm
\BotBlankLine
%

%% file: algs/enum-axp-cxp.tex
%
%
\SetKwFunction{tfunc}{{\sc XpEnum}} %
\Func \tfunc{$\fml{M}_{\pi}$,$\fml{C}$, $\pi$} \;

\Indp
\KwIn{
  $\fml{M}_{\pi}$: ML model,
  $\fml{C}$: Input cube,
  $\pi$: Prediction
}
\KwVars{
  $\fml{N}$ and $\fml{P}$ defined on the variables of $\fml{C}$
}
\TopBlankLine
{
  \lnlset{a4:01}{1}%
  $\fml{K}=(\fml{N},\fml{P})\gets(\emptyset,\emptyset)$ \tcp*[r]{Block AXps \& CXps}
  \lnlset{a4:02}{2}%
  \While{\true}{
    \lnlset{a4:03}{3}%
    $(\st_{\lambda},\lambda)\gets\FindMHS(\fml{P},\fml{N})$
    \tcp*[r]{MHS$\;\tn{of}\;\fml{P}\;\tn{st}\;\fml{N}$}
    \lnlset{a4:04}{4}%
    \lIf{$\neg\st_{\lambda}$}{\Break} 
    \lnlset{a4:05}{5}%
    $(\st_{\rho},\rho)\gets\SAT(\lambda\land\neg\fml{M}_{\pi})$\;

    \lnlset{a4:06}{6}%
    \uIf(\tcp*[f]{entailment holds}){$\neg\st_{\rho}$}{

      \lnlset{a4:07}{7}
      $\ReportAXp(\lambda)$\;

      \lnlset{a4:08}{8}
      $\fml{N}\gets\fml{N}\cup\NegLits(\lambda)$ \;
    }
    \lnlset{a4:9}{9}
    \uElse{%
      \lnlset{a4:10}{10}
      $\mu\gets\ExtractCXp(\fml{M}_{\pi}, \rho, \pi)$\;

      \lnlset{a4:11}{11}%
      $\ReportCXp(\mu)$\;

      \lnlset{a4:12}{12}%
      $\fml{P}\gets\fml{P}\cup\PosLits(\mu)$ \;
    }
  }
}
\Indm
\BotBlankLine
%